%% file: main.tex
\documentclass[10pt,twocolumn,letterpaper]{article}

\usepackage[pagenumbers]{cvpr} 

\input{preamble}
\definecolor{cvprblue}{rgb}{0.21,0.49,0.74}
\usepackage[pagebackref,breaklinks,colorlinks,allcolors=cvprblue]{hyperref}
\usepackage{multirow}
\def\confName{CVPR}
\def\confYear{2027}
\usepackage{url}

\title{POKE: Probe Offline, Keep Efficiently for Vision-Language-Action Models}

\author{
Jaehyuk Lee$^{1}$ \quad Yeajin Lee$^{2}$ \quad SuHyun Bae$^{1}$ \quad Donghun Lee$^{1}$ \thanks{Corresponding author.} \\
$^{1}$Department of Mathematics, Korea University, Seoul, Republic of Korea \\
$^{2}$Program in Actuarial Science and Financial Engineering, Korea University, Seoul, Republic of Korea \\
{\tt\small \{jaehyeokbear, gin1214, baeshstar, holy\}@korea.ac.kr}
}

\begin{document}
\maketitle
\input{sec/0_abstract}    
\input{sec/1_intro}
\input{sec/2_related_works}
\input{sec/3_method}
\input{sec/4_experiments_re}
\input{sec/5_discussion}
\section*{LLM Usage Statement}
We used Claude (Anthropic) and ChatGPT (OpenAI) for writing and coding
assistance. For writing, the tools were used throughout the main paper and the
supplementary material to proofread English grammar and style, and to condense
text to fit the page limit. They also gave feedback on clarity and organization,
such as which paragraphs to shorten or move to the appendix, and proposed
revised wording for paragraphs and figure captions, which the authors reviewed
and edited. For code, the tools were used to write plotting code for the
figures, to refactor our research code, and to port the public VLA-Cache and
VLA-Pruner implementations from the VLA-Pruner repository into our codebase as
baselines. The authors designed POKE, ran and verified all experiments, and
checked and cited all references themselves; the tools did not generate
experimental results or data. The authors take full responsibility for the
content of this paper.

{
    \small
    \bibliographystyle{ieeenat_fullname}
    \bibliography{main}
}

\newpage
\input{sec/X_suppl}

\end{document}

%% file: sec/0_abstract.tex
\begin{abstract}
Visual token pruning reduces the inference cost of Vision-Language-Action (VLA) models, but attention-based scores do not directly measure a token’s effect on the predicted action. We define action sensitivity as the magnitude of the first-order action change when a token is replaced by its camera’s mean token and estimate it for all tokens with a single backward pass. 
Computing it online is costly: scoring on the full (unpruned) graph needs an extra forward pass at every control step. Scoring on the pruned forward pass is cheaper but causes scoring-graph lock-in: removed tokens lose their direct paths through later layers and tend to receive lower
scores, making those locations less likely to return. On replayed observations, even with a random conditioning mask, pruned-graph scoring re-selects 59\% of the mask’s tokens, versus 6\% expected by chance; attention histories show a similar bias.
POKE (Probe Offline, Keep Efficiently) computes full-graph action sensitivities offline and distills them into a student with under 0.3M parameters. At inference, the student scores the current observation without backpropagation or previous masks, and score-guided diversity selection retains the tokens. Keeping 32 of 512 visual tokens on LIBERO, POKE achieves 95.2\% success on OpenVLA-OFT and 90.8\% on $\pi_{0.5}$ at $1.85\times$ and $1.26\times$ speedups, outperforming attention-, diversity-, and distillation-based pruners, with larger margins on LIBERO-Plus.

\end{abstract}

%% file: sec/1_intro.tex
\section{Introduction}
\label{sec:intro}





Vision-language-action (VLA) models convert camera images and language instructions into robot actions. Their visual inputs often contain hundreds of tokens, making visual token pruning a practical way to reduce inference cost. Removing tokens in early layers reduces computation in every subsequent layer, so a pruner has to decide which tokens to keep while preserving the policy's actions. 
Existing methods, however, use attention as a proxy for this decision, including attention received by visual tokens~\cite{chen2024image}, temporal aggregation of action-to-visual attention~\cite{liu2025bridging}, and attention combined with inter-frame similarity~\cite{xu2026vla}. These signals describe how tokens are attended to or reused. Yet the tokens a model attends to most need not be the ones that determine its actions, and none of these signals directly measures how changing a token changes the resulting action.

\begin{figure*}[t]
    \centering
    \begin{subfigure}[b]{0.47\linewidth}
        \centering
        \includegraphics[width=\linewidth]{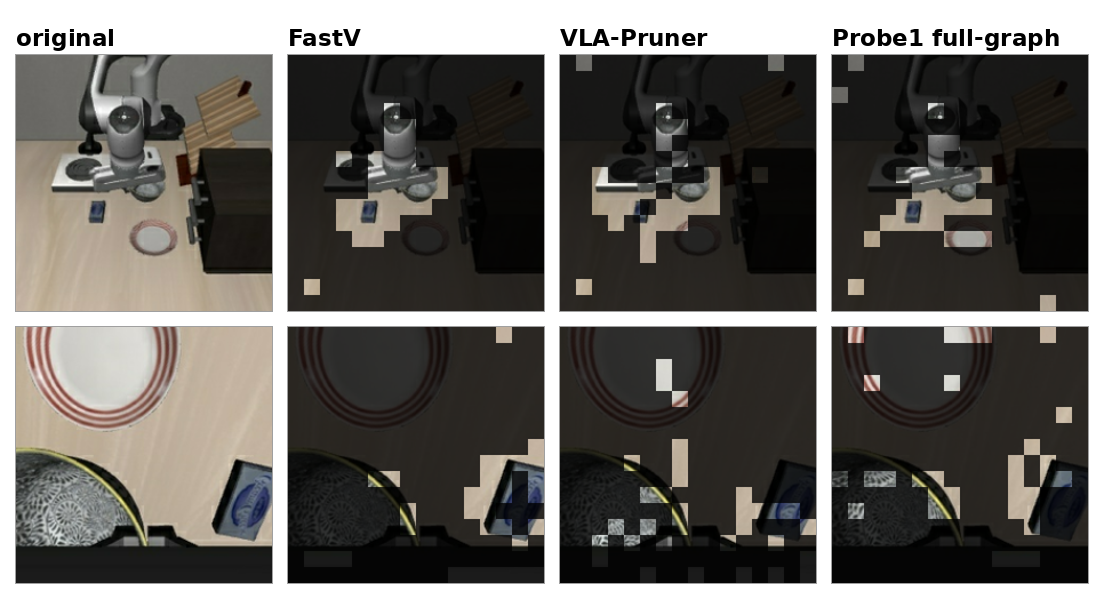}
        \caption{What to score: retained tokens before grasping}
        \label{fig:intro_score}
    \end{subfigure}
    \hfill
    \begin{subfigure}[b]{0.47\linewidth}
        \centering
        \includegraphics[width=\linewidth]{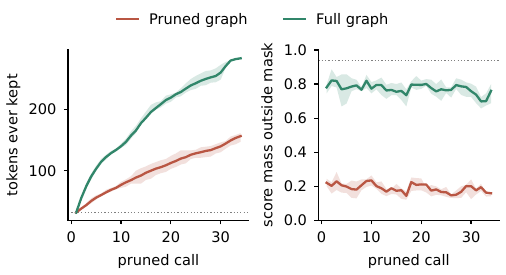}
        \caption{Where to score: online probes over control steps}
        \label{fig:intro_lockin}
    \end{subfigure}
    \caption{\textbf{Motivation} on OpenVLA-OFT. (a) LIBERO-Goal, 64/512 tokens; rows: third-person/wrist; clear patches: retained. Attention misses the target plate kept by action sensitivity. 
    (b) LIBERO-Long, $B=32$, 9 closed-loop episodes (mean, IQR). Left: cumulative number of distinct tokens ever retained; pruned-graph scoring keeps returning to the same tokens. Right: share of score assigned to tokens outside the previous mask; 19\% on the pruned graph vs. 77\% on the full graph (dotted: uniform scores).}
    
    \label{fig:intro_compare}
\end{figure*}

We therefore define token importance by action sensitivity: the magnitude of the first-order change in the action when a visual token is replaced with its camera-wise mean representation. Since importance is defined at the policy output, it applies to both parallel-regression and flow-based action heads, using the final action chunk for the latter. In Fig.~\ref{fig:intro_score}, attention-based masks keep only the bowl and gripper, whereas action sensitivity also keeps the target plate.


Computing exact sensitivities requires one backward pass per action dimension: 56 for a chunk of eight 7-D actions in OpenVLA-OFT. Projecting the action onto one shared random direction reduces this to a single backward pass for all tokens. Even so, scoring online on the full graph needs an extra unpruned forward and backward pass per control step, running $2.4\times$ and $3.1\times$ slower than the unpruned OpenVLA-OFT and $\pi_{0.5}$ policies, which defeats the purpose of pruning. 

A cheaper alternative scores tokens on the pruned forward pass and uses the scores to select the next frame’s tokens. Each frame enters with all visual tokens, but the previous mask removes some before the later layers, so those tokens lose their direct paths to the action and tend to score lower. When these scores set the next mask, removed locations are less likely to return. We call this scoring-graph lock-in (Fig.~\ref{fig:intro_lockin}).
Even when the previous mask is drawn at random, pruned-graph scores re-select 59\% of its tokens, against 6\% expected by chance, and switching to full-graph scoring raises closed-loop success (\S\ref{sec:exp_graph}).
The same lock-in appears in the attention histories that VLA pruners reuse from earlier control steps~\cite{liu2025bridging,wang2025specprune}. Scores therefore need a path independent of prior pruning decisions.

We propose POKE (Probe Offline, Keep Efficiently), which separates importance estimation from inference. POKE computes full-graph action-sensitivity labels offline on observations from the policy's fine-tuning demonstrations and distills them into a lightweight student with fewer than 0.3M parameters, keeping the policy frozen. Labeling takes under one GPU-hour per suite. At inference, the student predicts scores from the current visual tokens, language, and state context in a single forward pass, removing online backpropagation and any conditioning on previous masks. Tokens are then selected either by top-$k$ ranking or by diversity selection within a high-scoring candidate set; the latter reduces redundancy while keeping sensitivity-based candidate filtering. The full policy thus provides the training signal, and the deployed selector adds only a small scoring and selection cost.


On four LIBERO suites~\cite{liu2023libero}, keeping 32 of 512 visual tokens, POKE reaches 95.15\% success at $1.85\times$ on OpenVLA-OFT~\cite{kim2025fine} and 90.80\% at $1.26\times$ on $\pi_{0.5}$~\cite{intelligence2025pi}, exceeding VLA-Pruner by 5.90 and 15.75~pp, and the margin grows under the observation shifts of LIBERO-Plus (\S\ref{sec:exp_liberoplus}). In contrast, distilling attention labels instead of action sensitivity lowers success in all eight tested combinations (\S\ref{sec:exp_ablation}).

Our contributions are threefold:

\begin{itemize}

\item \textbf{Action sensitivity for visual token pruning.} An action-grounded, first-order measure of token importance estimated with one shared-probe backward pass, together with an exact pairwise analysis of its ranking behavior.

\item \textbf{Scoring-graph lock-in.} Controlled interventions and fixed-observation replays show that pruned-graph scores favor retained tokens even without feedback, for both gradients and attention histories.

\item \textbf{POKE via full-graph sensitivity distillation.} A compact selector trained offline that prunes without backpropagation or previous masks, improving the success-latency trade-off over attention-, diversity-, and distillation-based baselines on parallel-regression, flow-based, and autoregressive policies on LIBERO and SimplerEnv.

\end{itemize}

%% file: sec/2_related_works.tex
\section{Related Work}

\subsection{Visual Token Pruning}
\label{sec:related_visual_pruning}

Visual token reduction selects or aggregates tokens depending on the input. 
DynamicViT~\cite{rao2021dynamicvit} and EViT~\cite{liang2022not} prune or fuse less informative tokens.
In vision-language models (e.g., LLaVA~\cite{liu2024improved}), FastV~\cite{chen2024image} prunes visual tokens by early-layer attention, VisionZip~\cite{yang2025visionzip} combines encoder attention with similarity-based aggregation, and VisPruner~\cite{zhang2025beyond}, DART~\cite{wen2025stop}, and CDPruner~\cite{zhang2026beyond} add distinctive, non-redundant, or instruction-conditioned diverse tokens. \cite{zou2026don} argue that concentrating the budget on highly attended tokens discards holistic visual context. These works establish importance and diversity as complementary criteria; we study them for preserving actions in closed-loop.

\begin{figure*}[t]
  \centering
  \includegraphics[width=0.98\linewidth]{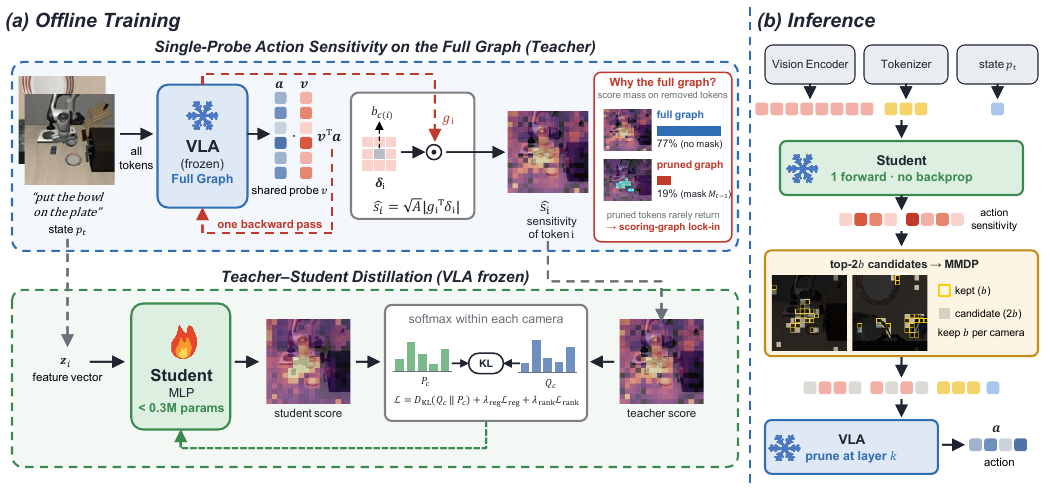}
  \caption{\textbf{Method overview. }(a) Offline training: a frozen full-graph VLA produces single-probe action-sensitivity labels, which are distilled into a lightweight student. (b) Inference: student predicts token scores, and MMDP selects diverse tokens from scoring candidates.}
  \label{fig:overview}
\end{figure*}

\subsection{Distillation-based Token Pruning}
\label{sec:related_distill}


Distillation transfers expensive importance signals to efficient token selectors. ToP~\cite{li2023constraint} distills final-layer token rankings into earlier layers, SDTP~\cite{tao2025saliency} trains in-network pruning modules on gradient-times-input saliency of hidden states with regression and ranking losses, and MAP~\cite{sun2026learning} distills middle-layer text-to-vision attention into an input-level predictor and combines it with diversity; we compare against an adaptation of MAP in \S\ref{sec:exp}. First-order expansions also underlie Taylor pruning criteria~\cite{molchanov2017pruning}, which estimate the change in loss. POKE instead distills the first-order change in the predicted robot action chunk under a camera-mean token replacement, scores tokens before the policy's Transformer layers, and examines how the computation graph used to obtain these scores affects selection over successive control steps.

\subsection{Efficient Visual Token Processing in VLAs}
\label{sec:related_vla}

VLA-specific methods exploit task context and temporal structure during control. Many select tokens from attention or task reused from earlier control steps: VLA-Pruner~\cite{liu2025bridging} fuses semantic importance with temporally aggregated action-to-visual attention, SpecPrune-VLA~\cite{wang2025specprune} combines previous- and current-frame cues, and ADP~\cite{pei2026action} adapts retention to manipulation stages. Others reduce computation more broadly: VLA-Cache~\cite{xu2026vla} reuses key--value states of visually stable tokens, SP-VLA~\cite{li2025spvla} and EfficientVLA~\cite{yang2026efficientvla} pair pruning with model scheduling or layer reduction and caching, and LightVLA~\cite{jiang2025better} learns differentiable selection during policy fine-tuning. POKE keeps the policy frozen and trains a separate selector on full-graph action-sensitivity labels; at deployment it uses only the current observation, without backpropagation or previous masks. 

%% file: sec/3_method.tex
\section{Method}
\label{sec:method}

Let $X=[x_1,\ldots,x_N]\in\mathbb R^{N\times d}$ denote the visual tokens at the projector output, $\ell$ the language input, and $p$ the proprioceptive state. The tokens come from $C$ cameras, with index set $\mathcal I_c$ for camera $c$ and camera assignment $c(i)$ for token $i$; we retain $B$ of them in total. The frozen policy outputs an action chunk $a_\theta(X,\ell,p)\in\mathbb R^A$, which for flow-based policies is the final chunk integrated from fixed initial noise; we omit $\ell$ and $p$ when unambiguous. A control step $t$ is one policy call on a new camera frame, with visual tokens $X_t$ and state $p_t$.
Figure~\ref{fig:overview} gives an overview.

\subsection{Action Sensitivity}
\label{sec:method_sens}


The most direct measure of a token's importance is how much the action changes when the token is removed. Actual deletion, however, changes the sequence length and the subsequent computation graph, so it admits no local derivative. We instead replace each token with the mean representation of its camera, which keeps the length fixed while suppressing token-specific information relative to the camera mean. 
\begin{equation}
    \begin{aligned}
     b_c &= \frac{1}{|\mathcal I_c|}\sum_{j\in\mathcal I_c}x_j, \qquad
     \delta_i = \operatorname{stopgrad}\bigl(x_i-b_{c(i)}\bigr).
    \end{aligned}
    \label{eq:delta}
\end{equation}
The baseline is computed from the original observation and held fixed along the replacement path $x_i(\alpha)=x_i-\alpha\delta_i$; camera-wise means avoid mixing representations from different views. A first-order expansion at $\alpha=0$ gives
\begin{equation}
    \begin{aligned}
     a_\theta(X^{i\leftarrow b})-a_\theta(X)&\approx-u_i,\\
     u_i:=J_i\delta_i\in\mathbb R^A,\qquad J_i&=\frac{\partial a_\theta}{\partial x_i}.
    \end{aligned}
    \label{eq:u}
\end{equation}
We define \textbf{action sensitivity} as $s_i=\|u_i\|_2$. Unlike a Jacobian norm, which reflects sensitivity to arbitrary input perturbations, $s_i$ measures the action change along the direction that pruning removes: the token's information beyond its camera mean. Because deployment deletes tokens after several layers rather than replacing them at the projector, $s_i$ remains a surrogate for deletion damage (\S\ref{sec:exp_surrogate}).

\subsection{Single-Probe Estimation}
\label{sec:method_probe}



Exact $u_i$ for all tokens needs $A$ backward passes ($A=56$ for OpenVLA-OFT, $A=70$ for $\pi_{0.5}$). 
Instead, we draw a shared direction $v$ uniformly from the unit sphere in $\mathbb R^A$; one backward pass through $v^\top a_\theta$ gives $g_i=J_i^\top v$ for every token, and
\begin{equation}
 \widehat s_i=\sqrt A\,|g_i^\top\delta_i|
             =\sqrt A\,|v^\top u_i|.
 \label{eq:probe}
\end{equation}
Since $\mathbb{E}[\mathbf v\mathbf v^\top]=\mathbf I/A$, $\hat s_i^2$ is unbiased for $s_i^2$, but its relative variance $2(A-1)/(A+2)$ approaches two (Appendix~\ref{app:surrogate}), so individual magnitudes are noisy.

Selection, however, depends only on within-camera rankings. 
Because $\mathbf v$ is shared, the probability of reversing a pair depends on their sensitivity ratio and the alignment of their action effects. For $s_i>s_j$, let $r=s_i/s_j$. The flip probability is $\frac{2}{\pi}\arctan(1/r)$ for orthogonal effects (0.20 at $r=3$, 0.06 at $r=10$), and zero for collinear effects.
Empirically, single-probe scores reach a within-camera Spearman correlation of 0.80 with exact sensitivities, and the 16 highest-sensitivity tokens per camera carry 84\% of the sensitivity energy (Appendix~\ref{app:ranking}).

\subsection{Full-Graph Teacher}
\label{sec:method_teacher}

\paragraph{Online scoring and lock-in.}


Action sensitivity needs a forward pass to $a_\theta$ followed by backpropagation. Scoring each new frame before pruning it would therefore require an extra unpruned forward pass at every control step, which defeats the purpose of pruning. A practical online probe instead scores within the pruned forward pass it already runs. At control step $t$, the policy receives a new frame $X_t$ with all visual tokens, prunes it with the mask $M_{t-1}$ chosen at the previous control step, and outputs the action; backpropagating through this same pruned forward gives the scores that select $M_t$ for the next frame:
\begin{equation}
 \widehat s^{\,\mathrm{pruned}}_{t,i}
 =\mathcal S_i(X_t,\ell,p_t;M_{t-1}),
 \label{eq:pruned_score}
\end{equation}
where $\mathcal S_i$ denotes Eq.~\eqref{eq:probe}, with probe randomness suppressed. 
Every frame arrives with all visual tokens, but tokens outside $M_{t-1}$ leave the scoring graph after the cut layer: they keep indirect gradient paths through earlier-layer attention and lose their direct paths through later layers. Scores read on the pruned graph therefore favor retained tokens, and because $M_t$ then prunes frame $t{+}1$, the bias carries over to the next control step. We call this scoring-graph lock-in.
Path deletion need not reduce gradient norms in general, so \S\ref{sec:exp_graph} directly quantifies the bias and isolates it from the feedback loop. Attention histories are mask-dependent in the same way, since excluded positions receive no new attention after the cut.

\paragraph{Offline full-graph labels.}
The teacher retains all visual tokens and computes
\begin{equation}
 s^T_{t,i}=\mathcal S_i(X_t,\ell,p_t;\mathbf 1).
 \label{eq:teacher}
\end{equation}
For a fixed observation and fixed randomness, these labels do not depend on a previous pruning mask. We compute and cache them offline on fine-tuning demonstration observations, keeping the policy frozen. 

For flow-based $\pi_{0.5}$, the action expert starts from noise $\varepsilon$ and updates an action chunk over $K$ Euler steps ($\tau: 1 \to 0$), conditioned on the visual prefix. We fix $\varepsilon$ and differentiate the final chunk:
\begin{equation}
 a_\theta(X,\ell,p)=\Phi_\varepsilon(X,\ell,p),
 \qquad J_i=\frac{\partial\Phi_\varepsilon}{\partial x_i}.
 \label{eq:flow}
\end{equation}
Backpropagation traverses the full integration, accounting for repeated use of the visual prefix and propagation through intermediate action states, so the label reflects the final action rather than an isolated velocity prediction.
Offline labeling uses one noise sample per observation, and the student, which does not receive $\varepsilon$, learns an observation-conditioned approximation; rankings vary far less across noise samples than across probes (Appendix~\ref{app:eps}). Appendix~\ref{app:policies} gives the autoregressive-action relaxation.

\newcommand{\dg}[1]{\textcolor{gray}{#1}}

\begin{table*}[t]
\centering
\caption{LIBERO success rates (\%) and speedups over Vanilla ($\times$) with $B$ retained visual tokens. $^\dagger$: our adaptation. Gray: online diagnostic probes (\S\ref{sec:exp_why}), excluded from bold along with Vanilla; on $\pi_{0.5}$ their backward passes run in eager mode, so their speedups are against eager Vanilla (Table~\ref{tab:app_lat_eager_pi05}). POKE uses its $B=32$ checkpoints at all budgets; timing details in Appendix~\ref{app:latency}.}
\label{tab:main}
\setlength{\tabcolsep}{4pt}
\resizebox{\textwidth}{!}{%
\begin{tabular}{cl cccccc c cccccc c cccccc}
\toprule
 & & \multicolumn{6}{c}{$B=128$} & & \multicolumn{6}{c}{$B=64$} & & \multicolumn{6}{c}{$B=32$} \\
\cmidrule{3-8}\cmidrule{10-15}\cmidrule{17-22}
 & Method & Spa. & Obj. & Goal & Long & Avg.($\uparrow$) & $\times$($\uparrow$) & & Spa. & Obj. & Goal & Long & Avg.($\uparrow$) & $\times$($\uparrow$) & & Spa. & Obj. & Goal & Long & Avg.($\uparrow$) & $\times$($\uparrow$) \\
\midrule
\multirow{13}{*}{\rotatebox[origin=c]{90}{OpenVLA-OFT}} & Vanilla & 98.6 & 97.2 & 97.6 & 93.4 & 96.70 & 1.00 & & 98.6 & 97.2 & 97.6 & 93.4 & 96.70 & 1.00 & & 98.6 & 97.2 & 97.6 & 93.4 & 96.70 & 1.00 \\
\cmidrule(l){2-22}
 & Random                 & 95.6 & 94.6 & 95.0 & 89.8 & 93.75 & 1.43 & & 72.4 & 80.8 & 52.4 & 33.8 & 59.85 & 1.78 & & 43.8 & 15.6 & 8.8 & 10.0 & 19.55 & 1.89 \\
 & FastV                  & 97.4 & 97.2 & 95.6 & 91.8 & 95.50 & 1.40 & & 95.8 & 94.6 & 93.6 & 84.2 & 92.05 & 1.73 & & 91.6 & 80.8 & 92.6 & 43.2 & 77.05 & 1.88 \\
 & DivPrune               & 98.0 & 96.4 & 93.6 & 90.2 & 94.55 & 1.37 & & 95.8 & 93.2 & 88.8 & 72.4 & 87.55 & 1.67 & & 76.4 & 90.4 & 80.8 & 40.8 & 72.10 & 1.84 \\
 & VLA-Cache              & 97.0 & \textbf{98.0} & 97.4 & 90.8 & 95.80 & 1.28 & & 96.0 & 92.4 & 94.2 & 80.2 & 90.70 & 1.44 & & 90.4 & 88.8 & 90.4 & 70.4 & 85.00 & 1.47 \\
 & VLA-Pruner             & 97.4 & \textbf{98.0} & \textbf{98.4} & 91.0 & 96.20 & 1.33 & & 96.0 & 97.8 & 97.6 & 88.6 & 95.00 & 1.66 & & 94.0 & 93.4 & 91.0 & 78.6 & 89.25 & 1.82 \\
 & SpecPrune-VLA$^\dagger$ & 97.6 & 97.0 & 96.2 & 89.4 & 95.05 & 1.35 & & 98.0 & 97.0 & 95.8 & 86.4 & 94.30 & 1.65 & & 96.6 & 92.4 & 86.8 & 72.6 & 87.10 & 1.80 \\
 & MAP$^\dagger$          & 98.6 & 97.2 & 96.4 & \textbf{92.0} & 96.05 & 1.36 & & 98.4 & 96.8 & \textbf{98.0} & 89.0 & 95.55 & 1.68 & & 97.4 & 95.6 & 94.4 & 86.4 & 93.45 & 1.83 \\
\cmidrule(l){2-22}
 & \dg{Probe top-$k$ (pruned)} & \dg{97.0} & \dg{96.8} & \dg{98.0} & \dg{90.8} & \dg{95.65} & \dg{0.84} & & \dg{95.4} & \dg{90.8} & \dg{94.6} & \dg{85.4} & \dg{91.55} & \dg{0.97} & & \dg{92.4} & \dg{70.8} & \dg{90.0} & \dg{59.6} & \dg{78.20} & \dg{1.05} \\
 & \dg{Probe MMDP (pruned)}    & \dg{97.4} & \dg{98.4} & \dg{98.2} & \dg{90.6} & \dg{96.15} & \dg{0.81} & & \dg{96.4} & \dg{98.0} & \dg{98.0} & \dg{90.2} & \dg{95.65} & \dg{0.97} & & \dg{96.2} & \dg{96.4} & \dg{91.6} & \dg{63.4} & \dg{86.90} & \dg{1.04} \\
 & \dg{Probe top-$k$ (full)}   & \dg{98.4} & \dg{98.4} & \dg{98.0} & \dg{91.2} & \dg{96.50} & \dg{0.28} & & \dg{96.8} & \dg{98.4} & \dg{98.0} & \dg{90.0} & \dg{95.80} & \dg{0.34} & & \dg{98.4} & \dg{97.0} & \dg{96.2} & \dg{81.8} & \dg{93.35} & \dg{0.41} \\
\cmidrule(l){2-22}
 & POKE (top-$k$)         & \textbf{99.2} & \textbf{98.0} & 98.0 & 91.0 & \textbf{96.55} & 1.42 & & 98.2 & \textbf{98.8} & \textbf{98.0} & \textbf{93.4} & \textbf{97.10} & 1.74 & & \textbf{98.2} & 96.8 & \textbf{97.4} & 82.8 & 93.80 & 1.89 \\
 & POKE (MMDP)            & 98.6 & \textbf{98.0} & 96.0 & 91.8 & 96.10 & 1.35 & & \textbf{99.0} & 98.0 & 96.0 & 92.8 & 96.45 & 1.68 & & 97.8 & \textbf{97.6} & 97.2 & \textbf{88.0} & \textbf{95.15} & 1.85 \\
\midrule
\multirow{12}{*}{\rotatebox[origin=c]{90}{$\pi_{0.5}$}} & Vanilla & 98.6 & 97.8 & 97.2 & 92.4 & 96.50 & 1.00 & & 98.6 & 97.8 & 97.2 & 92.4 & 96.50 & 1.00 & & 98.6 & 97.8 & 97.2 & 92.4 & 96.50 & 1.00 \\
\cmidrule(l){2-22}
 & Random                 & 73.4 & 82.2 & 80.2 & 68.4 & 76.05 & 1.22 & & 50.4 & 48.8 & 45.8 & 30.6 & 43.90 & 1.30 & & 11.6 & 20.4 & 33.6 & 4.8 & 17.60 & 1.31 \\
 & FastV                  & 95.8 & 97.6 & 95.6 & 90.4 & 94.85 & 1.22 & & 86.0 & 93.4 & 82.8 & 75.2 & 84.35 & 1.27 & & 40.6 & 74.0 & 54.2 & 20.8 & 47.40 & 1.29 \\
 & DivPrune               & 98.0 & 97.2 & 96.8 & 89.0 & 95.25 & 1.16 & & 94.2 & 97.4 & 94.0 & 84.6 & 92.55 & 1.22 & & 83.2 & 84.4 & 80.0 & 63.8 & 77.85 & 1.28 \\
 & VLA-Pruner             & \textbf{99.0} & 98.2 & 97.4 & \textbf{90.6} & \textbf{96.30} & 1.12 & & 91.8 & 94.2 & 92.0 & 85.2 & 90.80 & 1.20 & & 68.8 & 92.0 & 77.4 & 62.0 & 75.05 & 1.24 \\
 & SpecPrune-VLA$^\dagger$ & 89.4 & 95.2 & 84.8 & 73.8 & 85.80 & 1.15 & & 65.8 & 71.2 & 73.2 & 43.8 & 63.50 & 1.22 & & 26.8 & 42.0 & 51.8 & 9.4 & 32.50 & 1.26 \\
 & MAP$^\dagger$          & 98.8 & \textbf{98.6} & 97.4 & 90.0 & 96.20 & 1.14 & & 96.2 & 94.8 & 95.4 & 81.8 & 92.05 & 1.23 & & 91.0 & 94.8 & 91.2 & 68.4 & 86.35 & 1.26 \\
\cmidrule(l){2-22}
 & \dg{Probe top-$k$ (pruned)} & \dg{97.8} & \dg{97.8} & \dg{96.0} & \dg{90.4} & \dg{95.50} & \dg{0.48} & & \dg{93.8} & \dg{92.0} & \dg{91.6} & \dg{81.0} & \dg{89.60} & \dg{0.47} & & \dg{61.8} & \dg{76.8} & \dg{78.0} & \dg{46.6} & \dg{65.80} & \dg{0.48} \\
 & \dg{Probe MMDP (pruned)}    & \dg{98.0} & \dg{98.2} & \dg{96.6} & \dg{90.0} & \dg{95.70} & \dg{0.47} & & \dg{97.0} & \dg{95.6} & \dg{96.6} & \dg{87.4} & \dg{94.15} & \dg{0.47} & & \dg{83.2} & \dg{89.6} & \dg{88.6} & \dg{72.2} & \dg{83.40} & \dg{0.48} \\
 & \dg{Probe top-$k$ (full)}   & \dg{98.0} & \dg{98.6} & \dg{97.4} & \dg{91.2} & \dg{96.30} & \dg{0.32} & & \dg{97.2} & \dg{96.8} & \dg{95.8} & \dg{87.4} & \dg{94.30} & \dg{0.32} & & \dg{89.4} & \dg{93.4} & \dg{93.2} & \dg{79.4} & \dg{88.85} & \dg{0.32} \\
\cmidrule(l){2-22}
 & POKE (top-$k$)         & 97.8 & 98.0 & 96.6 & 89.6 & 95.50 & 1.22 & & 96.6 & 97.8 & 93.6 & \textbf{89.6} & 94.40 & 1.27 & & 84.6 & 86.2 & 77.0 & 68.4 & 79.05 & 1.29 \\
 & POKE (MMDP)            & 98.8 & 98.4 & \textbf{97.6} & 90.4 & \textbf{96.30} & 1.14 & & \textbf{96.8} & \textbf{98.0} & \textbf{97.6} & 89.0 & \textbf{95.35} & 1.22 & & \textbf{97.0} & \textbf{96.2} & \textbf{93.6} & \textbf{76.4} & \textbf{90.80} & 1.26 \\
\bottomrule
\end{tabular}}
\end{table*}

\subsection{Lightweight Student}
\label{sec:method_student}
\label{sec:method_loss}

The POKE student predicts within-camera importance from the current observation.
For each token, we combine a low-dimensional projection $h_i$ with camera and
scene means $\bar h_{c(i)},\bar h$, language features $h_\ell$, state $p$, spatial
and camera coordinates $q_i$, and magnitude $\nu_i=\log\operatorname{RMS}(x_i)$:
\begin{equation}
\begin{aligned}
 z_i&=[h_i;\bar h_{c(i)};\bar h;h_\ell;p;q_i;\nu_i],
    \widehat r_i=g_\phi(z_i).
\end{aligned}
\label{eq:student}
\end{equation}
The scoring head is a two-layer MLP with hidden width 64. Including the input
projections, the student has approximately 271K parameters for OpenVLA-OFT and
140K for $\pi_{0.5}$. Its inputs are available from projector outputs and input
embeddings, before the policy's Transformer layers.

Teacher scores are log-transformed and standardized within each camera to form
targets $y_c$. With $Q_c=\operatorname{softmax}(y_c)$ and
$P_c=\operatorname{softmax}(\widehat r_c)$, we minimize
\begin{equation}
\begin{aligned}
 \mathcal L &= \mathbb E_c\bigl[D_{\mathrm{KL}}(Q_c\,\|\,P_c)\bigr]\\
 &\quad +\lambda_{\mathrm{reg}}\mathcal L_{\mathrm{reg}}
          +\lambda_{\mathrm{rank}}\mathcal L_{\mathrm{rank}}.
\end{aligned}
\label{eq:loss}
\end{equation}
The KL, regression, and pairwise ranking terms supervise the importance distribution, standardized scores, and relative ordering, respectively; the latter two follow SDTP~\cite{tao2025saliency}.
Loss and training details are provided in Appendix~\ref{app:student_details}.
At deployment, the student produces all scores in one forward pass, without teacher evaluation, policy backpropagation, or previous-mask inputs, which removes the mask-conditioning path of Eq.~\eqref{eq:pruned_score}.

\subsection{Score-Guided Selection}
\label{sec:method_select}


We allocate $b=B/C$ tokens to each camera. \textbf{Top-$k$} retains the $b$ highest student scores. To reduce redundancy, \textbf{MMDP} (max--min diversity pruning) applies the max--min selection of DivPrune~\cite{alvar2025divprune}, which operates there on all visual tokens, only to a candidate set $\mathcal C_c$ of the top $\min(2b,|\mathcal I_c|)$ scores. With cosine distance $d(i,j)$ between projector embeddings $x_i$ and $x_j$, the first token is the candidate farthest from its nearest neighbor, and we then repeatedly add
\begin{equation}
 i^*=\arg\max_{i\in\mathcal C_c\setminus\mathcal S_c}
       \min_{j\in\mathcal S_c}d(i,j)
 \label{eq:mmdp}
\end{equation}
until $|\mathcal S_c|=b$. Scores thus set relevance and embedding distance sets redundancy~\cite{peng2005feature}. Selection costs $O(b^2d)$ per camera and is included in all reported latencies; unselected tokens are removed at the cut layer (\S\ref{sec:exp_setup}).

%% file: sec/4_experiments_re.tex
\section{Experiments}
\label{sec:exp}

\subsection{Experimental Setup}
\label{sec:exp_setup}


We evaluate OpenVLA-OFT~\cite{kim2025fine} (parallel regression, suite-specific checkpoints) and $\pi_{0.5}$~\cite{intelligence2025pi} (flow matching, one checkpoint for all suites) on the four LIBERO suites, with 500 episodes per suite and initial states matched across methods. Both policies receive 512 visual tokens from two cameras. We retain $B\in\{128,64,32\}$, split equally between cameras, and delete the rest at the input to layer 4 (OFT) or 3 ($\pi_{0.5}$); the first three control steps of each episode are unpruned for all methods, following VLA-Pruner’s history warmup, although POKE needs none. 
Speedups are paired median latencies per pruned control step against Vanilla, including scoring and selection; $\pi_{0.5}$ timings use CUDA graphs except for online probes, which are timed against eager Vanilla (Appendix~\ref{app:eval}).
Autoregressive OpenVLA on LIBERO and a $\pi_0$ re-implementation on SimplerEnv~\cite{li24simpler} are reported in Table~\ref{tab:openvla_main} (Appendices~\ref{app:openvla} and~\ref{app:simpler}). Robustness to observation shift is evaluated on LIBERO-Plus (\S\ref{sec:exp_liberoplus}).

All token-deletion methods share the budgets and cut. The cut and attention layers follow the released VLA-Pruner code~\cite{liu2025bridging} on OFT and its $\pi_0$ setting on $\pi_{0.5}$; $B=128$ and $64$ match its 25\% and 12.5\% retention, and $B=32$ extends it. We compare against \textbf{Random}, attention-based \textbf{FastV}~\cite{chen2024image}, diversity-based \textbf{DivPrune}~\cite{alvar2025divprune}, \textbf{VLA-Pruner (VP)} with temporal action-to-visual attention, and, on OFT, the key--value reuse of \textbf{VLA-Cache}~\cite{xu2026vla}. \textbf{SpecPrune-VLA$^\dagger$}~\cite{wang2025specprune} keeps the static stage of SpecPrune-VLA, which joins the previous control step's attention-selected tokens with early-layer attention and frame-difference tokens, at a fixed budget (Appendix~\ref{app:specprune}). \textbf{MAP$^\dagger$} adapts the attention-distillation pruner MAP~\cite{sun2026learning}, keeping its teacher, predictor, and selection rule but training with our objective (Appendix~\ref{app:map}). Training it with MAP’s original objective instead changes success by at most 2.0 pp. POKE students are trained once per policy--suite pair on full-graph $P=1$ labels from fine-tuning demonstrations (Appendix~\ref{app:student_details}). As diagnostics for lock-in, \textbf{online probes} score the mask for the next control step on the pruned graph or with an extra full-graph forward pass (\S\ref{sec:method_teacher}). Implementation details appear in Appendix~\ref{app:baselines}.

\begin{table}[t]
    \centering
    \caption{\textbf{Additional policies and benchmark} at $B=32$ of 256 visual
    tokens (success, \%). Top: autoregressive OpenVLA on LIBERO. Bottom:
    open-pi-zero $\pi_0$ (Bridge checkpoint) on SimplerEnv WidowX visual matching,
    240 episodes per task; its student is trained on real BridgeData V2 frames.
    Details and full results appear in Appendices~\ref{app:openvla}
    and~\ref{app:simpler}.}
    \label{tab:openvla_main}
    \resizebox{\linewidth}{!}{%
    \begin{tabular}{clccccc}
    \toprule
     & Method & Spatial & Object & Goal & Long & Avg.($\uparrow$) \\
    \midrule
    \multirow{6}{*}{\rotatebox[origin=c]{90}{OpenVLA}}
     & Vanilla        & 84.6 & 84.0 & 74.0 & 51.4 & 73.50 \\
     & FastV          & 61.2 & 74.2 & 58.8 & 36.4 & 57.65 \\
     & VLA-Cache      & 67.4 & 71.0 & 60.2 & 34.2 & 58.20 \\
     & VLA-Pruner     & 75.6 & 76.2 & 61.4 & 42.0 & 63.80 \\
     & POKE (top-$k$) & 72.8 & 81.2 & 70.0 & 44.2 & 67.05 \\
     & POKE (MMDP)    & \textbf{78.8} & \textbf{82.2} & \textbf{72.4} & \textbf{53.6} & \textbf{71.75} \\
    \midrule
     & Method & Carrot & Eggplant & Spoon & Stack & Avg.($\uparrow$) \\
    \midrule
    \multirow{5}{*}{\rotatebox[origin=c]{90}{$\pi_0$}}
     & Vanilla        & 57.1 & 84.6 & 82.1 & 49.2 & 68.23 \\
     & FastV          & 50.8 &  7.5 & 61.2 & 39.6 & 39.79 \\
     & VLA-Pruner     & \textbf{52.9} & 20.4 & 67.9 & 35.8 & 44.27 \\
     & POKE (top-$k$) & 52.1 & 20.0 & 61.7 & 43.3 & 44.27 \\
     & POKE (MMDP)    & 52.1 & \textbf{30.8} & \textbf{72.5} & \textbf{47.1} & \textbf{50.63} \\
    \bottomrule
    \end{tabular}}
\end{table}

\subsection{Main Results: Success and Runtime}
\label{sec:exp_main}
\label{sec:exp_latency}

At $B=32$, POKE (MMDP) achieves 95.15\% average success at $1.85\times$ speedup on OpenVLA-OFT and 90.80\% at $1.26\times$ on $\pi_{0.5}$ (Table~\ref{tab:main}), exceeding VP by 5.90 and 15.75~pp at comparable speeds. The learned selector adds no measurable cost: POKE (top-$k$) is within 0.2~ms of FastV on $\pi_{0.5}$ and no slower on OpenVLA-OFT, and MMDP adds 1.6--5.1~ms depending on the budget (Appendix~\ref{app:latency}). The smaller speedup on $\pi_{0.5}$ comes from the policy's latency profile: pruning cuts VLM-backbone FLOPs by $4.4\times$ and prefix latency by $2.46\times$, but the prefix is only 35\% of Vanilla latency (28.5 of 81.4~ms; Table~\ref{tab:app_lat_stage_pi05}), so no token pruner can exceed $1.54\times$ here.

\noindent\textbf{Against MAP$^\dagger$.} The strongest baseline at this budget is MAP$^\dagger$, which also combines distilled importance with diversity selection; POKE (MMDP) is higher on both policies (+1.7 and +4.45~pp). Without diversity, the comparison depends on label density. On OFT Long, POKE (top-$k$) reaches 82.8\% against 74.8\% for MAP with its original objective and top-$k$ selection (Appendix~\ref{app:map}). 
On $\pi_{0.5}$, POKE (top-$k$) with $P=1$ trails $\text{MAP}^\dagger$, whose rule includes diversity, while $P=8$ lifts top-$k$ above it on Long and Goal (75.4 vs. 68.4\% and 92.8 vs. 91.2\%; Table~\ref{tab:probe_count}).

\noindent\textbf{Budgets and other policies.} At $B=128$ the non-random methods are within 2~pp on OFT, where even Random retains 93.75\%. On $\pi_{0.5}$, $B=64$ is the more practical operating point: POKE (MMDP) stays within 1.15 pp of Vanilla at $1.22\times$. POKE also outperforms VP on autoregressive OpenVLA and SimplerEnv (Table~\ref{tab:openvla_main}). On SimplerEnv, the $\pi_0$ student is trained on real BridgeData V2 frames and evaluated in simulation; despite this real-to-sim gap, POKE (MMDP) retains 74\% of Vanilla success versus 65\% for VP.


\begin{figure*}[t]
    \centering
    \begin{subfigure}[b]{0.99\linewidth}
        \centering
        \includegraphics[width=\linewidth]{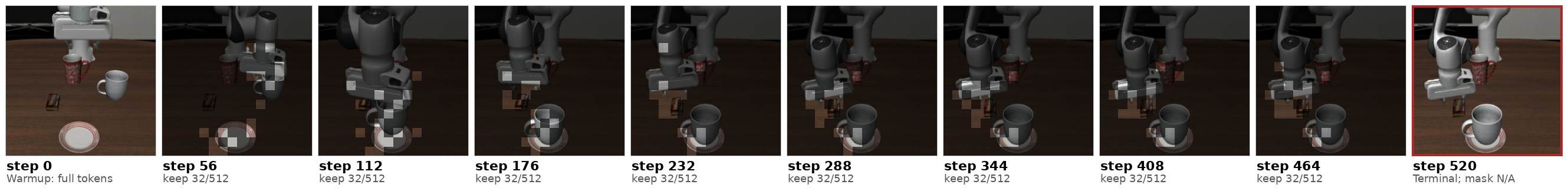}
        \caption{VLA-Pruner}
    \end{subfigure}
    \begin{subfigure}[b]{0.99\linewidth}
        \centering
        \includegraphics[width=\linewidth]{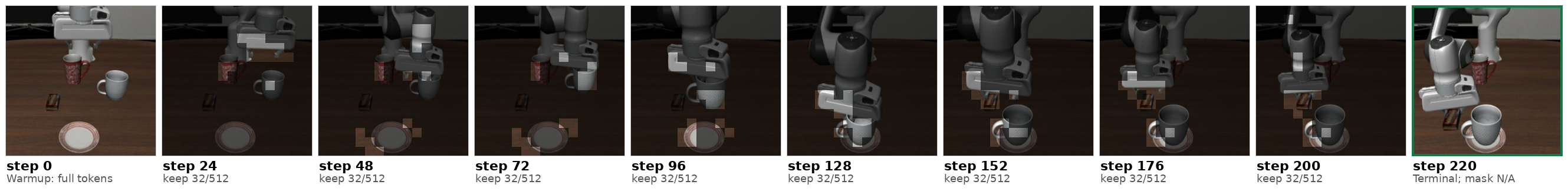}
        \caption{POKE (top-$k$)}
    \end{subfigure}
    \caption{\textbf{Qualitative rollouts on OpenVLA-OFT Long task 6} ($B=32$) from the
    same initial state, showing frames and the applied third-person mask. VLA-Pruner
    times out at step 520 (red); POKE (top-$k$) succeeds at step 220 (green).
    Figure~\ref{fig:app_rollout_full} adds more results.}
    \label{fig:accumul_compare}
\end{figure*}

\subsection{Robustness to Observation Shift}
\label{sec:exp_liberoplus}

\begin{table}[t]
    \centering
    \caption{\textbf{LIBERO-Plus retention by perturbation category} at $B=32$ on a
    stratified 25\% subset (2,508 perturbed tasks, one episode each), pooled over the
    four suites. Vanilla: success (\%); other rows: success relative to Vanilla.
    Policies and students are unchanged (trained on unperturbed demonstrations).
    Per-suite success appears in Appendix~\ref{app:liberoplus}.}
    \label{tab:liberoplus}
    \setlength{\tabcolsep}{3pt}
    \resizebox{\linewidth}{!}{%
    \begin{tabular}{clcccccccc}
    \toprule
     & Method & Camera & Robot & Lang. & Light & Backgr. & Noise & Layout & Total \\
    \midrule
    \multirow{7}{*}{\rotatebox[origin=c]{90}{\footnotesize OpenVLA-OFT}}
     & Vanilla (\%)   & 66.4 & 41.0 & 80.7 & 91.9 & 89.6 & 74.4 & 83.7 & 74.4 \\
     & FastV          & 0.56 & 0.51 & 0.74 & 0.63 & 0.73 & 0.58 & 0.66 & 0.64 \\
     & DivPrune       & 0.46 & 0.35 & 0.54 & 0.47 & 0.56 & 0.38 & 0.54 & 0.48 \\
     & VLA-Pruner     & 0.66 & 0.69 & 0.89 & 0.86 & 0.83 & 0.78 & 0.81 & 0.79 \\
     & MAP$^\dagger$  & 0.72 & 0.87 & 0.85 & 0.89 & 0.95 & 0.69 & 0.85 & 0.82 \\
     & POKE (top-$k$) & 0.82 & \textbf{0.89} & \textbf{0.97} & \textbf{0.98} & \textbf{0.96} & \textbf{0.93} & 0.92 & \textbf{0.92} \\
     & POKE (MMDP)    & \textbf{0.84} & 0.77 & 0.95 & 0.97 & 0.95 & 0.87 & \textbf{0.94} & 0.90 \\
    \midrule
    \multirow{7}{*}{\rotatebox[origin=c]{90}{$\pi_{0.5}$}}
     & Vanilla (\%)   & 71.4 & 74.8 & 84.3 & 94.0 & 93.7 & 87.1 & 86.1 & 83.6 \\
     & FastV          & 0.33 & 0.28 & 0.52 & 0.50 & 0.52 & 0.34 & 0.52 & 0.43 \\
     & DivPrune       & 0.38 & 0.33 & 0.56 & 0.66 & 0.66 & 0.38 & 0.69 & 0.52 \\
     & VLA-Pruner     & 0.30 & 0.33 & 0.54 & 0.73 & 0.63 & 0.33 & 0.61 & 0.49 \\
     & MAP$^\dagger$  & 0.35 & 0.35 & 0.57 & 0.55 & 0.58 & 0.36 & 0.55 & 0.47 \\
     & POKE (top-$k$) & 0.52 & 0.48 & 0.67 & 0.78 & 0.70 & 0.48 & 0.77 & 0.63 \\
     & POKE (MMDP)    & \textbf{0.72} & \textbf{0.72} & \textbf{0.90} & \textbf{0.97} & \textbf{0.91} & \textbf{0.64} & \textbf{0.89} & \textbf{0.82} \\
    \bottomrule
    \end{tabular}}
\end{table}

\begin{table}[t]
    \centering
    \caption{\textbf{Teacher supervision} at $B=32$ (success, \%). Only the teacher
    labels differ (Appendix~\ref{app:student_details}). $\pm$: mean$\pm$SD over three
    training seeds, separate from the Table~\ref{tab:main} checkpoint; other entries use the Table~\ref{tab:main} checkpoint.}
    \label{tab:teacher}
    \newcommand{\sd}[1]{{\scriptsize$\pm$#1}}
    \resizebox{\linewidth}{!}{%
    \begin{tabular}{llcccc}
    \toprule
     & & \multicolumn{2}{c}{OpenVLA-OFT} & \multicolumn{2}{c}{$\pi_{0.5}$} \\
    \cmidrule(lr){3-4}\cmidrule(lr){5-6}
    Suite & Teacher & Top-$k$ & MMDP & Top-$k$ & MMDP \\
    \midrule
    \multirow{2}{*}{Object} & Attention   & 93.4 & 96.0 & 82.8\sd{1.62} & 94.4\sd{2.02} \\
                            & Action sens. & \textbf{96.8} & \textbf{97.6} & \textbf{86.0}\sd{1.44} & \textbf{96.4}\sd{2.84} \\
    \midrule
    \multirow{2}{*}{Long}   & Attention   & 80.40\sd{2.38} & 83.20\sd{1.28} & 62.8\sd{0.48} & 71.8\sd{2.46} \\
                            & Action sens. & \textbf{83.93}\sd{0.81} & \textbf{87.53}\sd{1.86} & \textbf{68.8}\sd{0.98} & \textbf{76.4}\sd{1.12} \\
    \bottomrule
    \end{tabular}}
\end{table}

POKE students are trained only on unperturbed demonstrations, so we test whether their rankings transfer to LIBERO-Plus~\cite{fei2025liberoplus}, which perturbs each LIBERO task along seven dimensions: camera viewpoint, robot initial state, language, lighting, background, sensor noise, and object layout. All methods and both policies run the same stratified subset with the settings of \S\ref{sec:exp_setup}. Because Vanilla itself degrades, Table~\ref{tab:liberoplus} reports retention, success relative to Vanilla, for each perturbation category.

A POKE variant has the highest retention in every category on both policies, although on OpenVLA-OFT the margins over MAP$^\dagger$ are small for robot initial state (0.89 vs.\ 0.87) and background (0.96 vs.\ 0.95). Camera shifts move objects to different patch positions, so they test whether the student relies on a positional prior: POKE retains 82--84\% of Vanilla success on OpenVLA-OFT against 66\% for VP and 72\% for MAP$^\dagger$, and POKE (MMDP) retains 72\% on $\pi_{0.5}$ against at most 38\% for the baselines. Over all categories, POKE (MMDP) retains 82\% on $\pi_{0.5}$ against 43--52\%, and its 27.4~pp gap to VP exceeds the 15.75~pp gap on unperturbed LIBERO (Table~\ref{tab:libero_plus_suite}). MAP$^\dagger$, which exceeds VP by 11.3~pp on LIBERO with $\pi_{0.5}$, falls below it here (39.3 vs.\ 41.0\%). 

\begin{table}[t]
    \centering
    \caption{\textbf{History-graph interventions} (OFT Long, $B=32$, single runs). Pruned/full: graph used to compute history or probe scores; current-frame signals are computed before the cut; overlap: Jaccard.
    VP denotes VLA-Pruner; SpecPrune$^\dagger$ uses its fixed-budget static stage (Appendix~\ref{app:lockin}).}
    \label{tab:lockin}
    \setlength{\tabcolsep}{2pt}
    \resizebox{\linewidth}{!}{%
    \begin{tabular}{lcccccc}
    \toprule
     & & \multicolumn{2}{c}{Success rate} & &
     \multicolumn{2}{c}{Overlap} \\
    \cmidrule(lr){3-4}\cmidrule(lr){6-7}
    Method & Current-frame signal & pruned & full & & pruned & full \\
    \midrule
    VP: action attention only & -- & 3.4 & 69.4 & & 0.995 & 0.664 \\
    VP: combined & Semantic attention & 78.6 & 82.4 & & 0.515 & 0.364 \\
    \midrule
    SpecPrune$^\dagger$: history only & -- & 4.8 & 84.4 & & 1.000 & 0.431 \\
    SpecPrune$^\dagger$: combined & Local attention + frame difference
     & 72.6 & 80.6 & & 0.545 & 0.343 \\
    \midrule
    Action-sensitivity probe (top-$k$) & -- & 59.6 & 81.8 & & 0.85 & 0.41 \\
    \bottomrule
    \end{tabular}}
\end{table}

\subsection{Effect of Action-Sensitivity Supervision}
\label{sec:exp_ablation}
\label{sec:exp_teacher}
\label{sec:exp_need_score}


\noindent\textbf{Changing only the teacher.}
With the student, objective, and selection rule fixed, replacing action-sensitivity labels with full-graph action-to-visual attention, the signal behind VP's temporal score, 
lowers success in all eight policy–suite–rule combinations (Table~\ref{tab:teacher}). Under top-$k$, which compares the two score pools without any diversity correction, the drop is 3.2–6.0 pp; under MMDP it remains 1.6--4.6 pp. With denser $P=8$ labels, the top-$k$ gap on $\pi_{0.5}$ Long widens to 12.6 pp (75.4 vs. 62.8; Table~\ref{tab:probe_count}).


\noindent\textbf{Action fidelity at shared observations.}
Applying each method’s masks to the same Vanilla-rollout observations (OFT Long, 625 frames), both POKE variants yield lower action deviations than VP and FastV on all ten tasks ($p=0.002$; Table~\ref{tab:paired}); Figure~\ref{fig:accumul_compare} shows a closed-loop example.


\noindent\textbf{Is diversity alone sufficient?}
No. DivPrune applies the same max--min rule to all tokens without scores. It is the weakest non-random baseline on OFT and trails POKE (MMDP) by 12.95~pp on $\pi_{0.5}$ (Table~\ref{tab:main}). Scores alone are not always sufficient either: MMDP improves POKE over top-$k$ by 11.75~pp on $\pi_{0.5}$, 4.70 on OpenVLA, and 1.35 on OFT. 
Top-$k$ can waste budget on redundant high-sensitivity patches, while unfiltered diversity selects background.

\begin{table}[t]
\centering
    \caption{\textbf{Teacher probe count} on $\pi_{0.5}$ Long and Goal ($B=32$; success \%, 500 episodes). One checkpoint per $P$; $P=1$ is from Table~\ref{tab:main}. Label costs are GPU-hours; Long $P=3,5$ are scaled from measured $P=8$. Inference cost is unchanged.}
    \label{tab:probe_count}
    \small
    \resizebox{\linewidth}{!}{
    \begin{tabular}{lcccccc}
    \toprule
    & \multicolumn{3}{c}{Long} & \multicolumn{3}{c}{Goal} \\
    \cmidrule(lr){2-4} \cmidrule(lr){5-7}
    Teacher & GPU-h & Top-$k$ & MMDP & GPU-h & Top-$k$ & MMDP \\
    \midrule
    $P=1$ & 0.7 & 68.4 & 76.4 & 0.3 & 77.0 & 93.6 \\
    $P=3$ & 2.0 & 70.2 & 80.0 & 0.9 & 86.8 & 94.0 \\
    $P=5$ & 3.3 & 72.4 & 80.4 & 1.1 & 85.2 & 93.4 \\
    $P=8$ & 5.4 & \textbf{75.4} & \textbf{83.0} & 2.3 & \textbf{92.8} & \textbf{94.2} \\
    \midrule
    \textit{Online probe (full, top-$k$)} & -- & \textit{79.4} & -- & -- & \textit{93.2} & -- \\
    \bottomrule
    \end{tabular}}
\end{table}

\subsection{Why Distill a Full-Graph Teacher?}
\label{sec:exp_why}
\label{sec:exp_graph}
\label{sec:exp_online_cost}


\noindent\textbf{Full-graph.}
At $B=32$, switching the online top-$k$ probe from pruned to full-graph scoring raises OFT success from 90.0\% to 96.2\% on Goal and from 59.6\% to 81.8\% on Long. On $\pi_{0.5}$, average success rises from 65.80\% to 88.85\% (Table~\ref{tab:main}). Full-graph scoring is costly, however, running at $0.41\times$ and $0.32\times$ the unpruned speed, respectively. POKE moves this computation offline through distillation.


\noindent\textbf{Attention histories.}
Lock-in also appears in attention histories (Table~\ref{tab:lockin}). With history as the only signal, pruned-graph history freezes the mask (overlap $\geq$ 0.995) and success collapses to 3.4\% (VP action attention) and 4.8\% (SpecPrune-VLA$^\dagger$); full-graph history restores 69.4\% and 84.4\%. On $\pi_{0.5}$ Long, full history raises the complete SpecPrune-VLA$^\dagger$ from 9.4\% to 50.4\% (Appendix~\ref{app:lockin}). The complete VP changes less (78.6\% to 82.4\%) because its current-frame semantic route already re-admits tokens: under pruned history, 69\% of re-entering tokens come from that route alone.

\noindent\textbf{Is the feedback loop necessary?}
No; the pruned graph biases scores even without feedback. We replay probe scoring on OFT Long demonstration frames ($B=32$, 3,157 frames), using the same probe on full and pruned graphs. 
When the conditioning mask is drawn randomly at each frame, a scorer that ignored the mask would re-select 6\% of it by chance; pruned-graph scoring re-selects 59\%. On these frames, tokens outside the mask hold 94\% of the full-graph score mass but only 28\% of the pruned-graph mass. In closed loop, where the mask is the probe's own previous selection, the shares are 77\% and 19\% (Fig.~\ref{fig:intro_lockin}).

\subsection{Student Scope and Approximations}
\label{sec:exp_analysis}
\label{sec:exp_surrogate}
\label{sec:exp_probe}
\label{sec:exp_transfer}

\noindent\textbf{Probe count and label cost.}
With $P=1$, caching teacher labels costs 0.3--0.7 GPU-hours per suite (Appendix~\ref{app:student_details}), whereas exact labels would need $A$ backward passes per frame, about 47 GPU-hours for $\pi_{0.5}$ Long alone. We train students on multi-probe labels for $\pi_{0.5}$ Long and Goal (Table~\ref{tab:probe_count}). On Long, $P=8$ improves success by 7.0 pp (top-$k$) and 6.6 pp (MMDP) and narrows the top-$k$ gap to the online full-graph probe from 11.0 to 4.0 pp. On Goal, it raises top-$k$ by 15.8 pp, to within 0.4 pp of the online probe. The MMDP margin over top-$k$ shrinks from 16.6 to 1.4 pp on Goal but stays near 8 pp on Long, so denser labels replace the diversity gain only on some suites. The cost (5.4 and 2.3 GPU-hours) is paid once offline.

\noindent\textbf{Student scope.}
The student is not a fixed positional prior: it keeps most of Vanilla success under camera shifts (\S\ref{sec:exp_liberoplus}), and zero-shot cross-suite students still outperform VP (Spatial$\to$Long 84.4 vs.\ 78.6\%, Long$\to$Spatial 99.4 vs.\ 94.0\% with MMDP; Table~\ref{tab:app_cross_suite}).
For $\pi_{0.5}$, a single student trained jointly on all four suites matches per-suite students (90.55 vs.\ 90.80\% at $B=32$ with MMDP; Appendix~\ref{app:joint}).

\noindent\textbf{Surrogate fidelity.}
Replacement and deletion rank mask damage similarly on OFT Spatial (median within-frame Spearman 0.964). On OFT Goal demonstration frames, POKE (top-$k$) masks, with no backward pass, move the action about one quarter as far as random masks of the same budget (median ratio 0.254; Table~\ref{tab:surrogate}). Exact-sensitivity masks, which need 56 backward passes per frame, reach 0.250, and the single-probe teacher masks reach 0.285. The student's modest top-32 recall against its teacher (0.60; Appendix~\ref{app:student_details}) therefore does not turn into action error. The ranking is also meaningful in reverse: keeping the 32 tokens with the lowest exact sensitivity moves the action 1.84 times as far as a random mask (Table~\ref{tab:surrogate}).

%% file: sec/5_discussion.tex

\section{Discussion}
\label{sec:discussion}


\noindent\textbf{Lock-in is a property of the scoring graph.}
Our interventions (\S\ref{sec:exp_why}) suggest that methods reusing temporal signals need a mask-independent entry path or full-graph scoring to avoid lock-in.
Reusing a mask over several control steps causes a separate problem: the policy cannot attend to newly relevant regions until the mask is refreshed, and success decreases at longer reuse intervals in our tests (Table~\ref{tab:app_k}). 
Lock-in explains why online pruned-graph probing fails and why history-based pruners degrade at small budgets. It does not explain POKE's margin over per-frame selectors FastV and MAP$^\dagger$, which have no mask feedback; that margin comes from the supervision signal (\S\ref{sec:exp_ablation}).

\noindent\textbf{Why can the student match or exceed online probes?}
At $B=32$, the student matches or exceeds the online full-graph probe in average success under top-$k$ selection on OpenVLA-OFT and OpenVLA. We hypothesize two factors: first, the student scores the current observation directly rather than applying a mask chosen at the previous control step; second, training across frames averages the part of the single-probe error that is not predictable from the observation (Appendix~\ref{app:probe_var}).

On $\pi_{0.5}$, the student instead trails the online probe under top-$k$ (79.05\% vs.\ 88.85\% on average), and MMDP closes most of this gap (90.80\%). 
Single-probe variance is nearly identical for OpenVLA-OFT and $\pi_{0.5}$ (1.90 vs.\ 1.92; Appendix~\ref{app:probe_var}) and the online probe also uses $P=1$, so the shortfall is a distillation gap rather than a limit of the signal: $P=8$ labels largely close it (\S\ref{sec:exp_probe}).
Why $\pi_{0.5}$ students are more sensitive to label noise is untested; one candidate is that the action expert re-reads the same pruned prefix at all ten denoising steps.


\noindent\textbf{How many probes for the teacher?}
Our main results use $P=1$, the cheapest setting and the probe count of the online diagnostic; larger $P$ is an offline trade-off (\S\ref{sec:exp_probe}).


\noindent\textbf{Limitations.}
First, unlike training-free pruners, POKE requires offline teacher labeling and student training on demonstrations. Second, action sensitivity is a first-order, token-wise surrogate that does not explicitly model joint multi-token interactions, though candidate-restricted diversity selection helps mitigate redundancy in practice. Third, on $\pi_{0.5}$ the ten-step denoising loop is unaffected by token pruning and caps the end-to-end gain (\S\ref{sec:exp_main}). Finally, all closed-loop evaluations are in simulation; as partial evidence beyond it, the SimplerEnv student is trained on real BridgeData V2 frames, and POKE has the highest retention under the LIBERO-Plus camera, lighting, and noise shifts.

\section{Conclusion}
\label{sec:conclusion}

Action sensitivity measures which visual tokens a VLA policy needs to keep its actions, but computing it online on the pruned graph causes scoring-graph lock-in, a bias also observed in the pruned-graph attention histories of VLA-Pruner and SpecPrune-VLA. POKE computes it offline on the full graph and distills it into a small student that scores the current observation alone. Keeping 32 of 512 tokens, it reaches 95.2\% on OpenVLA-OFT, 90.8\% on $\pi_{0.5}$ at $1.85\times$ and $1.26\times$ speedups, ahead of attention-, diversity-, and distillation-based selectors and degrading less than VLA-Pruner on LIBERO-Plus; one student suffices for the jointly trained $\pi_{0.5}$. 
Real-robot evaluation and combining token selection with action-expert acceleration remain open.

%% file: sec/X_suppl.tex
\clearpage
\setcounter{page}{1}
\maketitlesupplementary

\newsavebox{\vlaAppendixTableBox}
\newcommand{\vlaAppendixFit}[2]{%
  \begingroup\small
  \sbox{\vlaAppendixTableBox}{#2}%
  \ifdim\wd\vlaAppendixTableBox>#1\relax
    \resizebox{#1}{!}{\usebox{\vlaAppendixTableBox}}%
  \else
    \usebox{\vlaAppendixTableBox}%
  \fi
  \endgroup
}

\section{Implementation and Reproducibility}
\label{app:impl}

\subsection{Policies and Pruning Paths}
\label{app:policies}

All policies use public checkpoints with frozen parameters. OpenVLA-OFT uses a 7B backbone with suite-specific checkpoints, 256 visual tokens from each of the third-person and wrist cameras, an additional token for the 8-dimensional proprioceptive state, and parallel prediction of eight 7-dimensional actions ($A=56$).
Autoregressive OpenVLA uses 256 tokens from one third-person camera and predicts seven action tokens ($A=7$). For OFT, sensitivities differentiate the action as the checkpoint returns it: the arm dimensions are unnormalized to execution units with the checkpoint's q01--q99 statistics, and the gripper dimension, which the checkpoint leaves unnormalized, stays in its $[-1,1]$ range.

\paragraph{$\pi_{0.5}$.}
We convert openpi's \texttt{pi05\_libero} checkpoint, jointly fine-tuned on all four LIBERO suites, to PyTorch bf16 (openpi commit \texttt{215abfb2}). The backbone is PaliGemma (Gemma 2B, 18 layers); layer $j$ of the 18-layer Gemma 300M action expert attends to the keys and values of prefix layer $j$. LIBERO uses agentview and wrist images from two of the three camera slots. We omit the masked, empty right-wrist slot and verify matching outputs in a self-test. SigLIP at 224-pixel resolution produces a $16\times16$ token grid per camera, for 512 visual tokens in total.

In this configuration, tokens are deleted at the input to layer 3: layers 0--2 process all tokens, and prefix/expert attention from layer 3 onward uses the retained visual tokens with their original position indices.
Actions are generated over a horizon of ten by ten Euler steps, with replanning every five environment steps. Sensitivities use the seven physical action dimensions, excluding 25 padding dimensions ($A=70$). We multiply normalized actions by their dimension-wise scales (quantile half-ranges), so the derivatives are expressed in execution units. The constant unnormalization offsets do not affect the Jacobian.

Episode limits are 220, 280, 300, and 520 environment steps for Spatial, Object, Goal, and Long. Reproduced Vanilla success rates are 98.6, 97.8, 97.2, and 92.4\% (average 96.50\%), compared with 98.8, 98.2, 98.0, and 92.4\% (96.85\%) reported for this checkpoint in the openpi repository. Each suite differs by at most 0.8~pp, within the sampling error of 500 episodes (about 0.8~pp at 97\% success). Rollout evaluation uses eager attention without compilation for deterministic execution; the separate CUDA-graph timing path is described in Appendix~\ref{app:latency_pi05}.

\paragraph{Discrete-action policies.}
For autoregressive OpenVLA, the executed argmax action is nondifferentiable. We fix the greedily decoded action-token sequence and use a teacher-forced forward pass to define a continuous surrogate,
\begin{equation}
 \widetilde a_k=\sum_m\operatorname{softmax}(z_k)_m\,\beta_m,
\end{equation}
where $\beta_m$ is the action-bin center at position $k$. Sensitivity is computed for this surrogate, rather than for the executed argmax mapping. Bin centres $\beta_m$ lie in the normalized range $[-1,1]$; as in \texttt{predict\_action}, arm dimensions are then unnormalized with the checkpoint's q01--q99 statistics and the gripper stays in $[-1,1]$, the same convention as for OFT.

\paragraph{Control steps and pruning location.}
Throughout, a control step (``call'') is one policy inference on a new camera frame, which executes an action chunk of several environment steps. Layer indices are 0-based. OFT deletes visual tokens at the input to layer 4; $\pi_{0.5}$ and OpenVLA use layer 3. Retained tokens keep their original position indices and thus their assigned positional phases. All nonvisual tokens, including BOS, proprioception, prompt, action slots, and stop tokens, are retained. OFT enforces a quota of $B/2$ per camera. The first three control steps of every episode are unpruned, so shorter episodes have a smaller fraction of pruned control steps (e.g., 91\% on Long and 77\% on Spatial in the recorded runs).

In OpenVLA, the mask chosen during prefill is held fixed throughout the seven action-token decoding steps. At each decoding step, the implementation checks that post-cut key--value caches have the expected shortened lengths and aborts otherwise. All reported rollouts pass this check. OFT runs use the bidirectional-attention fork required by its checkpoints. A first-frame check verifies nonzero upper-triangular attention at the cut layer, guarding against accidental use of the causal fork.

\subsection{Baseline Implementations}
\label{app:baselines}

\textbf{Random} samples $B$ tokens at each control step subject to camera quotas.
Our \textbf{VLA-Pruner} implementation combines the semantic signal at the cut with temporally smoothed action-to-visual attention from a deeper layer, and selects the mask within the current forward pass. Per camera, it takes the union of the top-$b$ tokens under each signal and applies the same greedy max--min routine as MMDP to this union. We port the selection logic while retaining OFT's bidirectional policy implementation, since replacing it with the repository's causal Transformers fork would change the policy. \textbf{FastV} ranks visual tokens by attention averaged across heads and queries at the same cut, retaining the top-$B/C$ per camera.
It omits the temporal signal and diversity filtering, providing a comparison at a matched pruning location.

\textbf{DivPrune}~\cite{alvar2025divprune} is run with our own implementation of its greedy max--min algorithm, the same routine that MMDP uses (Eq.~\eqref{eq:mmdp}), with the candidate set enlarged to all 256 tokens of a camera and no importance scores. As in DivPrune, distances are cosine distances between projector outputs, and the first token is the one farthest from its nearest neighbor. Two settings differ from the original. Selection is applied per camera with a quota of $B/2$, whereas DivPrune selects from the visual tokens of a single image. Tokens are deleted at the same cut as the other baselines (input to layer 4 for OFT and 3 for $\pi_{0.5}$), whereas DivPrune by default prunes before the language model; it also allows pruning at a decoder layer, and the later cut retains more information for the first layers (Appendix~\ref{app:surrogate}).

For $\pi_{0.5}$, FastV uses attention from prefix layer $k-1$. VP's $S_{\mathrm{vl}}$ uses that layer's attention, while $S_{\mathrm{act}}$ aggregates action-to-visual attention from the latter half of expert layers and flow steps, with history window 3 and $\gamma=0.8$. These settings follow the original $\pi_0$ configuration without modification.

\textbf{VLA-Cache} reuses key--value states across frames and is evaluated on OFT. We port its implementation from the public VLA-Pruner code base, which includes it as a baseline, without changing its reuse criterion. We did not implement it for $\pi_{0.5}$; caching may still apply to flow policies.

\textbf{Online probes} compute $P=1$ action sensitivities to select the mask for the \emph{next} control step, whereas the POKE student selects from the current observation. The pruned condition backpropagates through the current masked forward; the full condition adds a separate all-token forward for scoring.

\subsection{MAP Adaptation}
\label{app:map}

MAP~\cite{sun2026learning} was proposed for MLLMs. We reimplement MAP's Eqs.~(1)--(4) for OpenVLA-OFT from the paper, without an official release, and make the unspecified widths explicit below.

\paragraph{Teacher (QCTS).}
MAP labels use exactly the frames, episode split, and full-graph setting of the action-sensitivity cache (Appendix~\ref{app:student_details}), so both teachers are computed on identical observations. For each frame, the unpruned policy runs with SDPA attention, and we record attention at all 32 language model layers. Query rows are the question tokens of the OFT prompt, ``What action should the robot take to $\langle$instruction$\rangle$?'', excluding the \texttt{In:}/\texttt{Out:} markers, BOS, and action slots. Following MAP's Eq.~(1), attention to each visual token is averaged over heads and question tokens and then $\ell_1$-normalized jointly over all 512 visual tokens of both cameras, giving $\mathbf p^l$. The reference distribution $\mathbf p^l_{\mathrm{ref}}$ uses the same images and proprioceptive state with the question replaced by ``What is shown in this image?''. The teacher is $\mathbf p^{l^\star}$ with $l^\star=\arg\max_l D_{\mathrm{JS}}(\mathbf p^l,\mathbf p^l_{\mathrm{ref}})$ (MAP's Eq.~(2)), selected once per frame and shared by both cameras. Labels are computed from single frames, without temporal aggregation.

\paragraph{Predictor.}
Following MAP's Eq.~(3), question-token input embeddings and projector outputs pass through modality-specific two-layer MLPs ($4096\rightarrow256\rightarrow256$, GELU), followed by bias-free projections $\mathbf W_Q,\mathbf W_K$ to 64 dimensions. RoPE with the backbone's base frequency is applied at the original sequence positions of the question and visual tokens. The predicted distribution $\hat{\mathbf p}$ is the softmax of the scaled query--key products over the 512 visual tokens, averaged over question tokens. MAP does not specify these widths; our choice gives 2{,}261{,}504 parameters, about eight times the POKE student.

\paragraph{Objective.}
MAP minimizes $D_{\mathrm{KL}}(\mathbf p^{l^\star}\|\hat{\mathbf p})$ over all visual tokens. MAP$^\dagger$ instead uses our objective, so that the training loss matches POKE: the logarithms of $\hat{\mathbf p}$ and $\mathbf p^{l^\star}$ are standardized within each camera as in Eq.~\eqref{eq:app_norm}, and Eq.~\eqref{eq:app_loss} is applied with the same weights, pair sampling, and margin. Optimization settings and the episode split also match POKE, and the checkpoint with the lowest validation KL is selected.

Using MAP’s original objective instead changes success at B = 32 by at most 2.0 pp: 86.40 vs. 86.35\% (LIBERO) and 38.3 vs. 39.3\% (LIBERO-Plus) with $\pi_{0.5}$, and 84.4 vs. 86.4\% (LIBERO-Long) and 61.4 vs. 61.3\% (LIBERO-Plus) with OpenVLA-OFT. This is far below the gap to POKE, so the choice of objective does not explain the difference.

\paragraph{Selection and pruning.}
Selection is applied per camera with $b=B/2$. The top-$k$ variant retains the $b$ highest predicted scores. The diversity variant follows MAP's Eq.~(4): it initializes the retained set with the top $K_0=\lceil f b\rceil$ predicted tokens and then greedily adds $\arg\max_i \hat p_i\,\|\mathbf v_i-\mathbf B\mathbf B^{\top}\mathbf v_i\|_2$ over all remaining tokens of the camera, where $\mathbf B$ is an orthonormal basis of the retained projector outputs, maintained by Gram--Schmidt with reorthogonalization. We use $f=0.5$. Unlike MMDP, candidates are not restricted to the top-$2b$ scores, and importance multiplies the residual rather than gating eligibility. MAP$^\dagger$ in Table~\ref{tab:main} uses this rule, the stronger of the two: with MAP's original objective on OpenVLA-OFT LIBERO-Long at $B=32$, it reaches 84.4\% versus 74.8\% for top-$k$. MAP prunes before the first language-model layer; to match the other baselines, we instead delete tokens at the same cut (input to layer 4), with the same three unpruned warmup control steps. The predictor uses only projector outputs and prompt embeddings, so scoring still precedes the language model. MAP$^\dagger$ latency is measured with the same protocol as the other methods.

\paragraph{Adaptation to $\pi_{0.5}$.}
The $\pi_{0.5}$ variant follows the same procedure with three changes. The teacher compares all 18 PaliGemma prefix layers and selects, per frame, the layer that maximizes the JS divergence between attention under the task instruction and under the reference question. Query rows are the tokens of the task instruction itself, since the $\pi_{0.5}$ prompt has no surrounding question template, and keys are the 512 visual tokens of the two real cameras. The predictor takes 2048-dimensional projector outputs and instruction embeddings, with the same hidden width (256) and query--key width (64) as for OFT, giving about 1.2M parameters. Training uses our objective, selection uses MAP's Eq.~(4) with $f=0.5$, and tokens are deleted at the input to layer 3, as for the other $\pi_{0.5}$ baselines.

\subsection{SpecPrune-VLA Adaptation}
\label{app:specprune}

SpecPrune-VLA~\cite{wang2025specprune} combines action-level static pruning, layer-level dynamic pruning inside the language model, and an action-aware controller that switches between pruning profiles. The latter two change the number of retained tokens from call to call and from layer to layer. SpecPrune-VLA$^\dagger$ keeps only the static stage and fixes its output to $B$ tokens, deleted at the same cut as the other baselines (input to layer 4 for OFT and 3 for $\pi_{0.5}$) after the same three unpruned warmup calls. We port the selection from the public OpenVLA-OFT code.

\paragraph{Token sets.}
For each camera, $b=B/2$ is split as $K_{\mathrm{global}}:K_{\mathrm{local}}:K_{\mathrm{dyn}}=30:24:20$, the base values of SpecPrune-VLA, by largest remainder ($7:5:4$ at $B=32$). The token score is the attention from instruction tokens to each visual token, averaged over heads and instruction tokens, as in the released code; the fixed prompt prefix ``In: What action should the robot take to'' is excluded. The paper writes the score as visual-to-instruction attention; both exist under OFT's bidirectional attention, and we follow the code.
\begin{itemize}
    \item $V_{\mathrm{global}}$: the highest-scoring tokens at the 15th and the last
    layer (0-based 14 and 31) of the \emph{previous} call. The two layers alternate
    until $K_{\mathrm{global}}$ distinct tokens are chosen, which replaces the
    original union of per-layer top-$K/2$ sets, whose size varies.
    \item $V_{\mathrm{local}}$: the same score at layers 0 and 1 of the current call,
    before the cut, alternating in the same way.
    \item $V_{\mathrm{dyn}}$: the patches least similar to the frame six environment
    steps earlier, among those with cosine similarity below $\tau$ ($0.986$
    third-person, $0.98$ wrist). Similarity is computed on the $14\times14$ pixel
    patches aligned with the visual tokens, as in the released code. Six
    environment steps is the code's comparison interval when the controller is off.
\end{itemize}
The retained set is $V_{\mathrm{global}}\cup V_{\mathrm{local}}\cup V_{\mathrm{dyn}}$. When the sets overlap, the remaining slots are filled in $V_{\mathrm{global}}$ order; ties among tokens without a global score are broken by the current local score.

\paragraph{History graph and path conditions.}
As published, the previous call's attention comes from its pruned forward, so tokens deleted at that call receive no global score. The full-history condition of Table~\ref{tab:lockin} instead reads it from a separate unpruned forward on the same observation and discards that forward's actions. Path conditions disable $V_{\mathrm{local}}$ and $V_{\mathrm{dyn}}$; their share of $b$ is then filled in $V_{\mathrm{global}}$ order, so the budget is unchanged.

\paragraph{Adaptation to $\pi_{0.5}$.}
The $\pi_{0.5}$ variant differs in three settings. $V_{\mathrm{global}}$ uses prefix layers 8 and 17 of the 18 PaliGemma layers, the relative depth of the 15th and 32nd of 32. The query rows are all valid language tokens, since the $\pi_{0.5}$ prompt has no fixed template and also carries the discretized state. The client sends the frame six environment steps earlier with each call; $\pi_{0.5}$ replans every five environment steps and OFT every eight. $V_{\mathrm{local}}$ uses prefix layers 0 and 1, before the cut at layer 3.

\paragraph{Latency.}
Reading attention at the local, cut, and global layers is part of the method and is included in the measured latency.

\subsection{Determinism and Timing Protocol}
\label{app:eval}

Repeated Vanilla evaluations on the same machine and kernels produced identical episode outcomes on OFT Long and OpenVLA Goal/Object.
For $\pi_{0.5}$, initial noise is seeded by (suite, task, episode, control step); all 30 rerun episodes matched. Online probes use the same noise as the executed action, including the additional full-graph scoring pass. These checks cover only the tested configuration and hardware. All runs and timings use a single NVIDIA RTX A6000 (48\,GB) with driver 535.309.01 and CUDA 12.2.

On OFT Long task 5, all four online conditions encountered a simulator \texttt{QACC} instability warning at the same simulation time, $t=0.5480$. The warning recurred deterministically from the same initial state; we did not check whether it affected the subsequent trajectories identically.

Timing starts from a prepared observation and ends when CPU actions are returned, excluding environment stepping and image preparation. It includes student scoring, candidate construction, and diversity selection. Methods are executed on the same observation in randomized order within each repetition, and speedups are paired ratios against Vanilla. We report results from five independent processes with 50 repetitions each, with no concurrent workloads on the machine. 
Each process discards GPU warmup iterations before timing; this is separate from the three unpruned control steps at the start of each episode, which are not timed. Reported latencies are medians over pruned calls, pooled across processes, so speedups are per pruned control step rather than amortized over episodes.

\section{POKE Student Architecture and Training}
\label{app:student_details}

\subsection{Teacher Labels}

Teacher labels use each policy's fine-tuning demonstrations. OFT uses the LIBERO RLDS data, with preprocessing and state normalization matched to the evaluation pipeline and checkpoint statistics. Demonstrations are subsampled every eight environment steps. Training and validation episodes are disjoint, with splits stratified by instruction. Validation labels are used only for checkpoint selection.

For $\pi_{0.5}$, we split the openpi fine-tuning dataset (\texttt{physical-intelligence/libero}) by suite, use ten demonstrations per task, subsample every eight environment steps, and reserve 20\% for validation. 
Each frame receives one initial-noise sample, 
$P>1$ labels use $v_0,\dots,v_{P-1}$. Labels differentiate the full-graph action in Eq.~\eqref{eq:flow}, without averaging over multiple noise samples.

\begin{table*}[tbp]
\centering
    \caption{POKE student training records. Frame counts follow subsampling every eight
    environment steps. Recall measures overlap between the student's and teacher's
    top-$B$ sets on validation observations. Label-generation cost is GPU time spent
    building the teacher cache.}
    \label{tab:app_train}
    \vlaAppendixFit{\textwidth}{
    \begin{tabular}{llcccccc}
    \toprule
    Backbone & Suite & Train & Val & Best epoch & Val KL & Recall@32 & Labels (GPU-h) \\
    \midrule
    OFT & Spatial & 1{,}238 & 352 & 8 & 0.1467 & 0.556 & 0.3 \\
    OFT & Object & 1{,}498 & 375 & 5 & 0.1458 & 0.560 & 0.4 \\
    OFT & Goal & 1{,}337 & 350 & 5 & 0.1509 & 0.597 & 0.3 \\
    OFT & Long & 2{,}766 & 691 & 7 & 0.1445 & 0.568 & 0.7 \\
    \midrule
    $\pi_{0.5}$ & Spatial & 1{,}297 & 312 & 18 & 0.1341 & 0.553 & 0.3 \\
    $\pi_{0.5}$ & Object & 1{,}525 & 376 & 17 & 0.1281 & 0.585 & 0.4 \\
    $\pi_{0.5}$ & Goal & 1{,}266 & 317 & 20 & 0.1367 & 0.540 & 0.3 \\
    $\pi_{0.5}$ & Long & 2{,}750 & 678 & 19 & 0.1327 & 0.536 & 0.7 \\
    $\pi_{0.5}$ & All (joint) & 6{,}838 & 1{,}683 & 18 & 0.1323 & 0.554 & -- \\
    \midrule
    OpenVLA & Spatial & 1{,}238 & 350 & 6 & 0.1513 & 0.548 & 0.3 \\
    OpenVLA & Object & 1{,}498 & 375 & 12 & 0.1419 & 0.562 & 0.4 \\
    OpenVLA & Goal & 1{,}299 & 351 & 8 & 0.1432 & 0.558 & 0.3 \\
    OpenVLA & Long & 2{,}765 & 689 & 10 & 0.1486 & 0.557 & 0.7 \\
    \bottomrule
    \end{tabular}
}
\end{table*}

For OFT, the selected checkpoints occur at epochs 5--8, well before the end of training. Their validation Recall@32 values are modest (Table~\ref{tab:app_train}), yet action deviations under POKE-selected masks can approach those of teacher-selected masks (Appendix~\ref{app:surrogate}). One possible explanation is that sensitivity energy is concentrated in a few tokens, but these measurements do not isolate the effect of missed tokens.

\paragraph{Attention-teacher control.}
Table~\ref{tab:teacher} changes only the teacher labels. The attention teacher relabels exactly the frames of the action-sensitivity cache; visual features and episode splits are copied unchanged, and only the score is replaced. For each frame, the unpruned policy runs with SDPA attention, and we read layer-15 attention from the 56 action-slot query rows to all 512 visual tokens, averaged over heads and action queries. This is the action-to-visual signal underlying VP's $S_{\mathrm{act}}$ (Appendix~\ref{app:lockin}), computed on the full graph from a single frame, without temporal history. For $\pi_{0.5}$, the attention teacher relabels the frames of its action-sensitivity cache in the same way. It reuses the initial noise of each frame's action-sensitivity label, integrates all ten flow steps on the unpruned graph, and reads attention from the action-horizon query rows of action-expert layer 8 to the 512 visual tokens in the prefix keys of the same layer. The scores are averaged over heads, horizon queries, and all ten flow steps. Capturing attention leaves the sampled actions unchanged. The scores pass through the same per-camera normalization (Eq.~\eqref{eq:app_norm}), and the POKE student, objective (Eq.~\eqref{eq:app_loss}), optimization, and selection rules are unchanged. The $\pi_{0.5}$ Object/Long comparisons and OFT Long report means and SDs over three additional training seeds (2, 5, 10), drawn at random; OFT Object uses the seed-0 checkpoint of Table~\ref{tab:main}, which reports seed-0 results for all trained selectors.

\subsection{Architecture}

\paragraph{Input features.}
A visual token $x_i\in\mathbb R^d$ ($d=4096$, or 2048 for $\pi_{0.5}$) is projected to 32 dimensions as
\begin{equation}
 h_i=\operatorname{GELU}\bigl(W_v\operatorname{LN}(x_i)+b_v\bigr),
\end{equation}
with non-affine LayerNorm. Camera and scene context are
\begin{equation}
 \bar h_c=\frac{1}{|\mathcal I_c|}\sum_{j\in\mathcal I_c}h_j,
 \qquad \bar h=\frac1C\sum_c\bar h_c.
\end{equation}
Language features $h_\ell$ are formed by averaging valid prompt input embeddings and applying a separate LayerNorm--linear--GELU projection to 32 dimensions; no deep contextual language states are required.

With $q_i$ containing two normalized patch coordinates and a camera identifier, and $\nu_i=\log\operatorname{RMS}(x_i)$, the scorer is
\begin{equation}
\begin{aligned}
 z_i&=[h_i;\bar h_{c(i)};\bar h;h_\ell;p;q_i;\nu_i],\\
 \widehat r_i&=W_2\operatorname{GELU}(W_1z_i+b_1)+b_2.
\end{aligned}
\end{equation}
The MLP hidden width is 64. Including both input projections, the parameter counts are 271,297 for OFT ($|p|=8$), 140,225 for $\pi_{0.5}$ ($d=2048$, $|p|=8$), and 270,785 for OpenVLA ($|p|=0$). Scores are unconstrained since selection uses within-camera rankings.

\subsection{Objective and Optimization}

\paragraph{Teacher normalization.}
For each observation, scores are normalized independently within each camera:
\begin{equation}
\label{eq:app_norm}
\begin{aligned}
 \widetilde\ell_i&=\log\max(\widehat s_i,10^{-12}),\\
 y_i&=\frac{\widetilde\ell_i-\mu_{c(i)}}{\max(\sigma_{c(i)},0.1)}.
\end{aligned}
\end{equation}
Here $\mu_c$ and $\sigma_c$ are the population mean and standard deviation across the camera's log scores. This transformation reduces scale variation and preserves ordering above the floor; scores clipped at the floor can tie.

\paragraph{Objective.}
With $T=1$, $Q_c=\operatorname{softmax}(y_c/T)$, and
$P_c=\operatorname{softmax}(\widehat r_c/T)$, the loss is
\begin{equation}
\label{eq:app_loss}
\begin{aligned}
 \mathcal L&=T^2\mathbb E_c[D_{\mathrm{KL}}(Q_c\|P_c)]\\
 &\quad+0.1\mathcal L_{\mathrm{SmoothL1}}(\widehat r,y)
       +0.2\mathcal L_{\mathrm{rank}}.
\end{aligned}
\end{equation}
The ranking term is
\begin{equation}
\begin{aligned}
 \mathcal L_{\mathrm{rank}}
   &=\mathbb E_{(i,j)}[\operatorname{softplus}(-d_{ij})],\\
 d_{ij}&=\operatorname{sign}(y_i-y_j)(\widehat r_i-\widehat r_j).
\end{aligned}
\end{equation}
Each iteration samples 256 within-camera index pairs $(i,j)$, shared across the batch and applied separately in each camera; pairs with $|y_i-y_j|\le0.1$ are excluded.

\paragraph{Optimization.}
We train for 20 epochs with AdamW (learning rate $3\times10^{-4}$, weight decay $10^{-4}$), batch size 4, and gradient clipping at 1.0. The checkpoint with the lowest validation KL is selected; the policy remains frozen.

\subsection{Probe Count for $\pi_{0.5}$ Teacher Labels}
\label{app:probe_count}

Table~\ref{tab:probe_count} evaluates students trained with different teacher probe counts on $\pi_{0.5}$ Long. A single run computes eight probes per frame (seeds 0--7) using the frames, episode split, and initial noise of the $P=1$ cache. The label for probe count $P$ uses the first $P$ probes:
\begin{equation}
 \widehat s_i^{(P)}=
 \Bigl(\frac1P\sum_{k=1}^{P}\widehat s_{i,k}^{\,2}\Bigr)^{1/2}.
\end{equation}
Its square averages $P$ unbiased estimates of $s_i^2$ (Appendix~\ref{app:probe_var}). Probe 0 reproduces the cached $P=1$ labels, with zero maximum relative difference on the 50 checked frames.

The student architecture, objective, 20 training epochs, and checkpoint selection by validation KL follow Appendix~\ref{app:student_details}. Rollouts use the 500 initial states and initial-noise seeds of Table~\ref{tab:main}. The eight-probe run takes 5.62\,s per frame on one RTX A6000 (5.4 GPU-hours for 3{,}428 frames), against 0.7 GPU-hours for the $P=1$ cache (Table~\ref{tab:app_train}).

\begin{table}[tbp]
    \centering
    \caption{Training records of $\pi_{0.5}$ Long students for teacher probe count $P$. Val KL and Recall@32 are measured against each row's own validation labels. Because the targets differ with $P$, these values do not directly compare student quality across rows. The $P=1$ row is from Table~\ref{tab:app_train}.}
    \label{tab:app_probe_count}
    \setlength{\tabcolsep}{6pt}
    \vlaAppendixFit{\linewidth}{%
    \begin{tabular}{lccc}
    \toprule
    Teacher & Best epoch & Val KL & Recall@32 \\
    \midrule
    $P=1$ & 19 & 0.1327 & 0.536 \\
    $P=3$ & 20 & 0.1032 & 0.583 \\
    $P=5$ & 20 & 0.0963 & 0.595 \\
    $P=8$ & 17 & 0.0886 & 0.611 \\
    \bottomrule
    \end{tabular}}
\end{table}

For $P=3$ and $P=5$, validation KL is lowest at the final epoch, whereas for $P=8$ it is lowest at epoch 17 and remains within 0.0886--0.0897 thereafter. The $P=3$ and $P=5$ students may therefore benefit from longer training, which we did not test. Because Table~\ref{tab:app_probe_count} uses a different validation target for each probe count, our downstream comparison uses closed-loop success (Table~\ref{tab:probe_count}).

\section{Additional Success-Rate Results}
\label{app:success}

\subsection{Task-Level Success}

\begin{table*}[tbp]
    \centering
    \caption{Task-level success rates (\%) for online methods and DivPrune on OpenVLA-OFT ($B=32$). The four methods are evaluated in one session per suite. Graph effect is full minus pruned scoring with top-k selection, in percentage points.}
    \label{tab:app_task_online}
    \vlaAppendixFit{\textwidth}{
    \begin{tabular}{lcccccccccc c}
    \toprule
    Method & 0 & 1 & 2 & 3 & 4 & 5 & 6 & 7 & 8 & 9 & Overall \\
    \midrule
    \multicolumn{12}{l}{\textit{Goal}} \\
    Probe top-k (pruned) & 96 & 98 & 96 & 72 & 96 & 88 & 70 & 100 & 94 & 90 & 90.0 \\
    Probe MMDP (pruned)  & 98 & 92 & 94 & 80 & 96 & 80 & 82 & 100 & 100 & 94 & 91.6 \\
    Probe top-k (full)   & 100 & 94 & 96 & 94 & 98 & 92 & 90 & 100 & 98 & 100 & 96.2 \\
    DivPrune             & 92 & 92 & 90 & 48 & 98 & 16 & 82 & 100 & 98 & 92 & 80.8 \\
    Graph effect & $+4$ & $-4$ & 0 & $+22$ & $+2$ & $+4$ & $+20$ & 0 & $+4$ & $+10$ & $+6.2$ \\
    \midrule
    \multicolumn{12}{l}{\textit{Long}} \\
    Probe top-k (pruned) & 54 & 78 & 80 & 82 & 36 & 58 & 30 & 82 & 28 & 68 & 59.6 \\
    Probe MMDP (pruned)  & 78 & 94 & 84 & 70 & 42 & 54 & 24 & 96 & 30 & 62 & 63.4 \\
    Probe top-k (full)   & 80 & 92 & 78 & 96 & 66 & 78 & 72 & 96 & 74 & 86 & 81.8 \\
    DivPrune             & 70 & 92 & 66 & 28 & 6 & 12 & 8 & 80 & 2 & 44 & 40.8 \\
    Graph effect & $+26$ & $+14$ & $-2$ & $+14$ & $+30$ & $+20$ & $+42$ & $+14$ & $+46$ & $+18$ & $+22.2$ \\
    \bottomrule
    \end{tabular}
    }
\end{table*}

On Long, full-graph top-k improves over pruned-graph top-k on nine of ten tasks. The largest gains occur on tasks 6 and 8 ($+42$ and $+46$ pp), where pruned scoring reaches only 30\% and 28\%. Task 7, already at 82\%, gains 14 pp. The graph effect is therefore broadly distributed across tasks, unlike the concentrated diversity gains below.

\subsection{Task-Level Contribution of Diversity}
\label{app:mmdp_task}

\begin{table*}[tbp]
    \centering
    \caption{MMDP minus top-k success rates (pp) for POKE on
    OpenVLA-OFT Long ($B=32$), shown separately for three training seeds.}
    \label{tab:app_mmdp_task}
    \vlaAppendixFit{\textwidth}{
    \begin{tabular}{lccccccccccc}
    \toprule
    Task & 0 & 1 & 2 & 3 & 4 & 5 & 6 & 7 & 8 & 9 & Overall \\
    \midrule
    Seed 2 & $+2$ & $+8$ & $+28$ & $+4$ & $+8$ & $-2$ & $+6$ & $+6$ & $-8$ & $-4$ & $+4.8$ \\
    Seed 5 & 0 & $+4$ & $+30$ & $-6$ & $+8$ & $-4$ & $+2$ & $+2$ & $-8$ & 0 & $+2.8$ \\
    Seed 10 & 0 & $+4$ & $+24$ & $-8$ & $+20$ & $-10$ & $+2$ & 0 & $-2$ & $+2$ & $+3.2$ \\
    \midrule
    Mean & $+0.7$ & $+5.3$ & $\mathbf{+27.3}$ & $-3.3$ & $+12.0$ & $-5.3$ & $+3.3$ & $+2.7$ & $-6.0$ & $-0.7$ & $+3.6$ \\
    \bottomrule
    \end{tabular}
    }
\end{table*}

Task 2 accounts for much of the gain ($+28/+30/+24$ pp across seeds), and task 4 improves consistently. Tasks 5 and 8 regress for every seed. Excluding task 2, average gains fall to $+2.22/-0.22/+0.89$ pp. Thus, diversity repeatedly recovers particular weak tasks rather than uniformly improving the suite.

\begin{table*}[t]
    \centering
    \caption{Task-level POKE success rates (\%) on $\pi_{0.5}$ at $B=32$. Each task is evaluated over 50 episodes. Top-$k$ and MMDP denote the two POKE selection rules. Entries report single-checkpoint results, rather than averages over training seeds; task IDs are suite-specific.}
    \label{tab:app_task_pi05}
    \setlength{\tabcolsep}{5pt}
    \small
    \begin{tabular}{c cc cc cc cc}
    \toprule
    & \multicolumn{2}{c}{Spatial}
    & \multicolumn{2}{c}{Object}
    & \multicolumn{2}{c}{Goal}
    & \multicolumn{2}{c}{Long} \\
    \cmidrule(lr){2-3}
    \cmidrule(lr){4-5}
    \cmidrule(lr){6-7}
    \cmidrule(lr){8-9}
    Task
    & Top-$k$ & MMDP
    & Top-$k$ & MMDP
    & Top-$k$ & MMDP
    & Top-$k$ & MMDP \\
    \midrule
    0 &  88 & 100 & 92 &  94 &  92 &  92 & 66 & 72 \\
    1 &  86 &  98 & 74 & 100 &  88 &  98 & 66 & 88 \\
    2 & 100 & 100 & 86 & 100 &  94 &  98 & 74 & 84 \\
    3 &  92 & 100 & 88 &  94 &  50 &  84 & 84 & 90 \\
    4 &  60 &  88 & 88 &  90 &  96 &  98 & 60 & 58 \\
    5 &  76 &  98 & 98 &  96 &  48 &  82 & 76 & 86 \\
    6 & 100 &  98 & 88 &  98 &  80 &  98 & 66 & 64 \\
    7 &  78 &  96 & 88 &  94 & 100 & 100 & 66 & 98 \\
    8 &  96 & 100 & 96 & 100 &  48 &  88 & 50 & 44 \\
    9 &  70 &  92 & 64 &  96 &  74 &  98 & 76 & 80 \\
    \midrule
    Mean
    & 84.6 & 97.0
    & 86.2 & 96.2
    & 77.0 & 93.6
    & 68.4 & 76.4 \\
    $\Delta$ (pp)
    & \multicolumn{2}{c}{+12.4}
    & \multicolumn{2}{c}{+10.0}
    & \multicolumn{2}{c}{+16.6}
    & \multicolumn{2}{c}{+8.0} \\
    \bottomrule
    \end{tabular}
\end{table*}

Table~\ref{tab:app_task_pi05} reports single-checkpoint results on $\pi_{0.5}$. MMDP improves success on 32 of 40 tasks, ties on three, and reduces success on five, yielding an average gain of 11.75 percentage points across the four suites.

\subsection{Full Qualitative Rollouts}
\label{app:qualitative_full}

\begin{figure*}[t]
    \centering
    \begin{subfigure}[b]{0.99\linewidth}
        \centering
        \includegraphics[width=\linewidth]{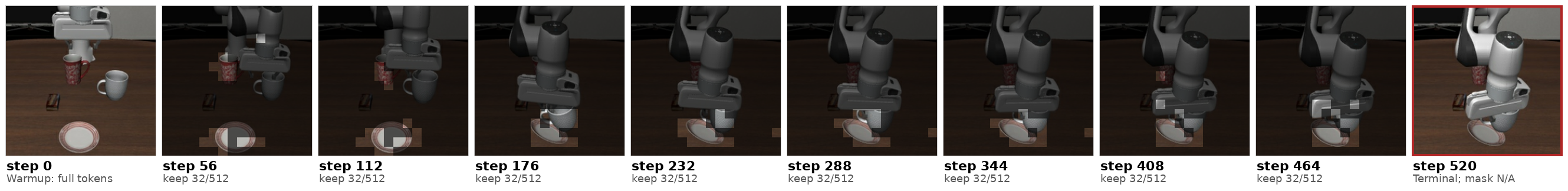}
        \caption{FastV}
    \end{subfigure}
    \begin{subfigure}[b]{0.99\linewidth}
        \centering
        \includegraphics[width=\linewidth]{result_figures2/applied_mask_cam0_vlapruner.png}
        \caption{VLA-Pruner}
    \end{subfigure}
    \begin{subfigure}[b]{0.99\linewidth}
        \centering
        \includegraphics[width=\linewidth]{result_figures2/applied_mask_cam0_student_topk.png}
        \caption{POKE (top-$k$)}
    \end{subfigure}
    \begin{subfigure}[b]{0.99\linewidth}
        \centering
        \includegraphics[width=\linewidth]{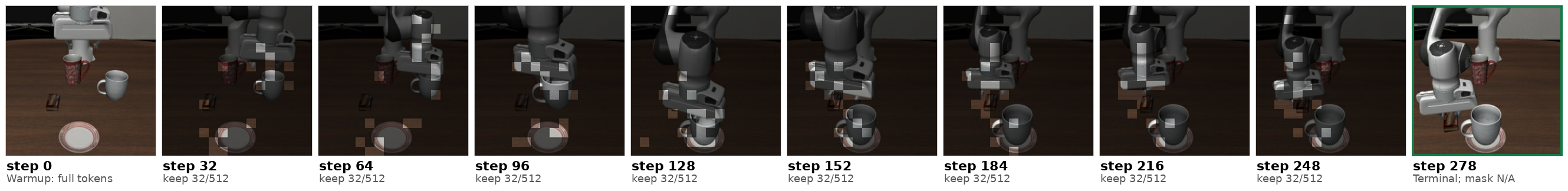}
        \caption{POKE (MMDP)}
    \end{subfigure}
    \caption{\textbf{All methods on OpenVLA-OFT Long task 6} ($B=32$), from the same
    initial state as Figure~\ref{fig:accumul_compare}. Layout and borders follow
    Figure~\ref{fig:accumul_compare}.}
    \label{fig:app_rollout_full}
\end{figure*}

Figure~\ref{fig:app_rollout_full} extends Figure~\ref{fig:accumul_compare} to all four methods on the same episode. FastV and VLA-Pruner reach the limit of 520 environment steps, while POKE (top-$k$) and POKE (MMDP) succeed at environment steps 220 and 278, respectively.

\subsection{Scoring Graphs and Mask Overlap}
\label{app:graph}

Table~\ref{tab:main} reports the corresponding success and runtime; Tables~\ref{tab:graph} and~\ref{tab:graph_pi05} add consecutive-mask overlap. Lower overlap alone does not explain higher success: on OFT, DivPrune has the lowest overlap and the lowest success.

\begin{table}[tbp]
    \centering
    \caption{Scoring graphs and mask overlap on OpenVLA-OFT ($B=32$).
    Overlap is consecutive-mask Jaccard, averaged first within episodes and then
    across episodes. Online methods are evaluated in one session per suite.
    POKE Long success is mean$\pm$SD over three training seeds; POKE overlap
    uses one checkpoint (seed 0). These seed averages differ from the
    single-checkpoint results in Table~\ref{tab:main}.}
    \label{tab:graph}
    \vlaAppendixFit{\linewidth}{
    \begin{tabular}{lcccc}
    \toprule
     & \multicolumn{2}{c}{Goal} & \multicolumn{2}{c}{Long} \\
    \cmidrule(lr){2-3}\cmidrule(lr){4-5}
    Method & Success & overlap & Success & overlap \\
    \midrule
    Probe top-k (pruned) & 90.0 & 0.81 & 59.6 & 0.85 \\
    Probe MMDP (pruned)  & 91.6 & 0.62 & 63.4 & 0.65 \\
    Probe top-k (full)   & 96.2 & 0.36 & 81.8 & 0.41 \\
    DivPrune             & 80.8 & 0.25 & 40.8 & 0.32 \\
    \midrule
    POKE (top-$k$)        & \textbf{97.4} & 0.53 & 83.93$\pm$0.81 & 0.62 \\
    POKE (MMDP)         & 97.2 & 0.39 & \textbf{87.53}$\pm$1.86 & 0.44 \\
    \bottomrule
    \end{tabular}
    }
\end{table}

\begin{table}[tbp]
    \centering
    \caption{Scoring graphs and mask overlap on $\pi_{0.5}$ ($B=32$), using the definition
    in Table~\ref{tab:graph}. Masks were recorded for Spatial and Object.
    All values use a single run, matching the success rates in Table~\ref{tab:main}.}
    \label{tab:graph_pi05}
    \vlaAppendixFit{\linewidth}{
    \begin{tabular}{lcccc}
    \toprule
     & \multicolumn{2}{c}{Spatial} & \multicolumn{2}{c}{Object} \\
    \cmidrule(lr){2-3}\cmidrule(lr){4-5}
    Method & Success & overlap & Success & overlap \\
    \midrule
    Probe top-k (pruned) & 61.8 & 0.68 & 76.8 & 0.69 \\
    Probe MMDP (pruned)  & 83.2 & 0.54 & 89.6 & 0.54 \\
    Probe top-k (full)   & 89.4 & 0.42 & 93.4 & 0.47 \\
    \midrule
    POKE (top-$k$)        & 84.6 & 0.74 & 86.2 & 0.78 \\
    POKE (MMDP)         & \textbf{97.0} & 0.55 & \textbf{96.2} & 0.54 \\
    \bottomrule
    \end{tabular}
    }
\end{table}

\subsection{History-Graph Interventions}
\label{app:lockin}

The attention rows of Table~\ref{tab:lockin} vary the history graph and selection rule on OFT Long. All conditions use $B=32$ (16 tokens per camera), deletion at the input to layer 4, $S_{\mathrm{act}}$ from layer 15, history window 3, $\gamma=0.8$, and three warmup control steps. We evaluate 500 matched initial states once per condition. Executed actions always come from the masked forward; $S_{\mathrm{vl}}$ is read from that forward in all conditions.

\paragraph{History graph.}
Only the source of new $S_{\mathrm{act}}$ history entries changes. The \emph{pruned} condition averages layer-15 attention across heads and action-slot queries, scatters it to retained token positions, and fills excluded positions with zero.
The \emph{full} condition adds an unpruned forward on the same observation to obtain all 512 entries and discards its actions. Full history is a mechanistic intervention, not an acceleration method.

\paragraph{Selection.}
VP forms the union of the top-16 $S_{\mathrm{vl}}$ and top-16 $S_{\mathrm{act}}$ candidates per camera, then selects 16 by max--min diversity. The $S_{\mathrm{act}}$-only condition selects its top-16 per camera with index-based tie breaking, matching the implementation's fallback selection.

\paragraph{Metrics.}
We count only pruned control steps and exclude overlap/re-entry transitions whose preceding control step is warmup. With retained sets $K_t,K_{t-1}$ of size $B$,
\begin{equation}
 J_t=\frac{|K_t\cap K_{t-1}|}{|K_t\cup K_{t-1}|}.
\end{equation}
We average $J_t$ first within episodes and then across episodes, as in Table~\ref{tab:graph}. Re-entry tokens are $K_t\setminus K_{t-1}$. Using the actual decision scores, we reconstruct the per-camera candidate sets $C_{\mathrm{vl}}$ and $C_{\mathrm{act}}$. A re-entry belongs to the $S_{\mathrm{vl}}$-only category if it lies in $C_{\mathrm{vl}}\setminus C_{\mathrm{act}}$.
The reported fraction pools such token counts over all eligible control steps, then divides by the pooled number of re-entry tokens. It is therefore weighted by re-entry counts, unlike the episode-averaged Jaccard statistic. For $S_{\mathrm{act}}$-only selection, $K_t=C_{\mathrm{act}}$, so this fraction is zero by definition and displayed as a dash. Act.\ support is the number of visual tokens with nonzero history-aggregated $S_{\mathrm{act}}$ at decision time. We discard each episode's first three pruned control steps and average support over all remaining control steps pooled across the 500 episodes, rather than averaging episode means. This removes the full-width warmup entries from the three-step history.

\begin{table*}[tbp]
    \centering
    \caption{Detailed history-graph interventions (OpenVLA-OFT Long, $B=32$, one run).
    Success is in percent. Trimmed overlap excludes each episode's first three
    pruned control steps, during which full-width warmup entries can remain in the
    history. Re-entry is the mean number of entering tokens per control step;
    $S_{\mathrm{vl}}$-only is the pooled re-entry fraction defined in the text.
    Act.\ support is the mean number of tokens with nonzero temporal scores
    over the trimmed interval.}
    \label{tab:app_lockin}
    \setlength{\tabcolsep}{3pt}
    \vlaAppendixFit{\textwidth}{%
    \begin{tabular}{llcccccc}
    \toprule
    Selection & History & Success & Overlap & Trimmed overlap & Re-entry & $S_{\mathrm{vl}}$-only & Act.\ support \\
    \midrule
    VLA-Pruner       & pruned & 78.6 & 0.515 & 0.518 & 10.0 & 0.69 & 49.02 \\
    VLA-Pruner       & full   & 82.4 & 0.364 & 0.368 & 14.6 & 0.48 & 512.00 \\
    S$_\mathrm{act}$ & pruned &  3.4 & 0.995 & 1.000 & 0.08 & --   & 32.13 \\
    S$_\mathrm{act}$ & full   & 69.4 & 0.664 & 0.671 & 6.2  & --   & 512.00 \\
    \bottomrule
    \end{tabular}}
\end{table*}

After omitting the first three pruned control steps, $S_{\mathrm{act}}$-only with pruned history has Jaccard 1.000 at the reported precision and mean temporal-score support 32.13, nearly equal to the retention budget ($B=32$). These observations are consistent with an effectively fixed mask. Pruned-history VP has a larger mean support of 49.02; full history gives support 512 at every retained measurement for both selection rules. Under pruned history, the temporal-score support is contained in the union of masks represented in the window. Outside that union, scores are zero: such tokens cannot outrank positive-score candidates, although zero-score ties can permit selection if too few positive candidates remain.

For pruned-history VP, the remaining 31\% of re-entries also belong to $C_{\mathrm{act}}$. After warmup history expires, nonzero temporal scores can recover tokens absent from the immediately preceding mask but present earlier in the window. The $S_{\mathrm{vl}}$ route provides candidates outside that support; membership in either candidate set is not a causal attribution of the final diversity selection.

\paragraph{Reproduction checks.}
Pruned-history VP matches the 78.6\% Long success rate in Table~\ref{tab:main}. The runner rejects mismatched cut, attention layer, history window, or budget. Matching aggregate success does not show that the masks or action trajectories are identical.

\paragraph{Comparisons across scoring methods.}
On OFT Long at $B=32$, the full-graph probe reaches 81.8\%, compared with 69.4\% for full-history $S_{\mathrm{act}}$ alone. Both use per-camera top-k and scores delayed by one control step, but the attention condition additionally averages a three-step history, and the probe runs were collected separately. Full-history VP reaches 82.4\%, compared with 88.0\% for POKE (MMDP). These comparisons do not isolate the scoring signal: temporal aggregation, selection rules, and mask timing differ.

\paragraph{SpecPrune-VLA$^\dagger$.}
The SpecPrune-VLA$^\dagger$ rows of Table~\ref{tab:lockin} use the settings of Appendix~\ref{app:specprune} on the same 500 OFT Long initial states, once per condition. Only the graph of the global attention changes between pruned and full history; $V_{\mathrm{local}}$ and $V_{\mathrm{dyn}}$ are always computed at the current control step before the cut. Overlap and re-entry exclude transitions from warmup control steps, without further trimming. A re-entering token is attributed to the set or sets that selected it at control step $t$; tokens added by the fill follow $V_{\mathrm{global}}$ order and are counted with $V_{\mathrm{global}}$. Shares pool token counts over all eligible control steps, as for VP.

\begin{table*}[tbp]
    \centering
    \caption{SpecPrune-VLA$^\dagger$ history-graph and path conditions (LIBERO-Long,
    $B=32$, 500 episodes, one run per condition). Re-entry is the mean number of
    entering tokens per pruned control step. The last four columns split re-entering
    tokens by the set that selected them: $V_{\mathrm{global}}$ order (including the
    fill), $V_{\mathrm{local}}$ only, $V_{\mathrm{dyn}}$ only, or both current-frame
    sets. Current-frame only is the sum of the last three, the analogue of the
    $S_{\mathrm{vl}}$-only fraction of Table~\ref{tab:app_lockin}.}
    \label{tab:app_lockin_sp}
    \setlength{\tabcolsep}{3pt}
    \vlaAppendixFit{\textwidth}{%
    \begin{tabular}{llccccccc}
    \toprule
    Paths & History & Success & Overlap & Re-entry & Current-frame only &
    $V_{\mathrm{global}}$ order & $V_{\mathrm{local}}$ & $V_{\mathrm{dyn}}$ / both \\
    \midrule
    \multicolumn{9}{l}{\textit{OpenVLA-OFT}} \\
    All (original)        & pruned & 72.6 & 0.545 &  8.89 & 1.00 & 0.00 & 0.44 & 0.54 / 0.02 \\
    All                   & full   & 80.6 & 0.343 & 14.93 & 0.61 & 0.39 & 0.26 & 0.34 / 0.01 \\
    $V_{\mathrm{global}}$ & pruned &  4.8 & 1.000 &  0.00 & --   & --   & --   & -- \\
    $V_{\mathrm{global}}$ & full   & 84.4 & 0.431 & 12.25 & 0.00 & 1.00 & --   & -- \\
    \midrule
    \multicolumn{9}{l}{$\pi_{0.5}$} \\
    All (original)        & pruned &  9.4 & 0.763 &  4.40 & 1.00 & 0.00 & 0.16 & 0.84 / 0.00 \\
    All                   & full   & 50.4 & 0.494 & 10.81 & 0.56 & 0.44 & 0.12 & 0.44 / 0.00 \\
    $V_{\mathrm{global}}$ & pruned &  5.0 & 1.000 &  0.00 & --   & --   & --   & -- \\
    $V_{\mathrm{global}}$ & full   & 68.2 & 0.539 &  9.63 & 0.00 & 1.00 & --   & -- \\
    \bottomrule
    \end{tabular}}
\end{table*}

With $V_{\mathrm{global}}$ alone and pruned history, the mask cannot change after warmup: tokens deleted at the previous control step have no global score, and the previously retained tokens fill all $b$ slots, so the overlap is 1.000 by construction. The same signal with full history reaches 84.4\%, above the original combination with either history. Under pruned history, all re-entries come from the current-frame sets, mostly $V_{\mathrm{dyn}}$; under full history, $V_{\mathrm{global}}$ order supplies 39\% of them, and adding the current-frame sets lowers success from 84.4\% to 80.6\%. The current-frame paths therefore compensate for the pruned history rather than adding information that the full-graph history lacks.

On $\pi_{0.5}$ the same pattern is stronger. Pruned-history SpecPrune-VLA$^\dagger$ is more persistent than on OFT (overlap 0.763 versus 0.545); its re-entries come mostly from $V_{\mathrm{dyn}}$ (84\%) and rarely from $V_{\mathrm{local}}$ (16\%, against 44\% on OFT), and its success (9.4\%) is close to Random (4.8\%). With $V_{\mathrm{global}}$ alone the mask again freezes (5.0\%). Full history raises $V_{\mathrm{global}}$ alone to 68.2\% and the original combination to 50.4\%, with layers, query tokens, budget, and initial noise unchanged, so the low Table~\ref{tab:main} result reflects the pruned history rather than the layer mapping of Appendix~\ref{app:specprune}. With full history, adding the current-frame sets lowers success by 17.8~pp, against 3.8~pp on OFT.

\subsection{Shadow Scoring on a Shared Trajectory}
\label{app:shadow}

Closed-loop comparisons of scoring graphs change the observations as well as the masks. To hold observations fixed, we replay $P=1$ probe scoring sequentially along the OFT Long fine-tuning demonstrations used for student training ($B=32$, 3{,}157 frames). No action is executed, so the observations do not depend on any condition. At each frame $t$, the same probe $v$ is applied to four graphs. \emph{Full} keeps all tokens; its per-camera top-16, $F_t$, is the reference selection. The reference is therefore the $P=1$
full-graph score with the same probe, not the exact Jacobian.
\emph{Pruned-self} deletes tokens outside the mask $M$ that this condition selected at the previous frame, reproducing the online pruned probe without its effect on the trajectory. \emph{Pruned-indep} uses a mask redrawn at every frame, independently of scores and previous selections, by sampling 16 of the 256 tokens of each camera uniformly without replacement. The graph is reduced, but no selection feeds back. \emph{Pruned-good} uses the full-graph probe's selection at the previous frame, a good mask that this condition did not choose. Each condition then selects its per-camera top-16, $K_t$. Frames are spaced every eight demonstration steps, matching the eight environment steps executed between OFT control steps. As in deployment, the first three frames of each episode serve as unpruned warmup and are excluded from all metrics, leaving 3{,}157 frames.

With conditioning mask $M$, top-$B$ recall is $|K_t\cap F_t|/B$ and mask overlap is $|K_t\cap M|/|K_t|$, with $|K_t|=32$. Excluded mass is the share of the condition's scores $s$ on tokens outside $M$; full-graph mass is the share of full-graph scores on the same tokens. Re-entry recall is $|K_t\cap(F_t\setminus M)|/|F_t\setminus M|$, averaged over frames with a nonzero denominator; the pooled ratio of total hits to total denominators gives nearly the same values (0.083, 0.078, and 0.168 for self, good, and indep). Spearman correlations with full-graph scores use all tokens. Values are means over frames.

\begin{table}[tbp]
    \centering
    \caption{Shadow scoring on OFT Long demonstration frames ($B=32$, 3{,}157
    frames). All conditions use the same probe at each frame and differ only in
    the graph; the reference is the $P=1$ full-graph selection.
    Excl.\ mass and Full mass are the shares of the condition's and the full-graph
    scores on tokens outside the conditioning mask.}
    \label{tab:app_shadow}
    \setlength{\tabcolsep}{3pt}
    \vlaAppendixFit{\linewidth}{%
    \begin{tabular}{lcccccc}
    \toprule
    Graph & Top-$B$ rec.\,$\uparrow$ & Overlap & Excl.\ mass & Full mass & Re-entry rec.\,$\uparrow$ & Spearman \\
    \midrule
    Full          & 1.000 & --    & --    & --    & --    & 1.000 \\
    Pruned-self   & 0.318 & 0.829 & 0.198 & 0.784 & 0.083 & 0.498 \\
    Pruned-good   & 0.395 & 0.857 & 0.186 & 0.717 & 0.077 & 0.516 \\
    Pruned-indep  & 0.218 & 0.594 & 0.284 & 0.937 & 0.168 & 0.447 \\
    \bottomrule
    \end{tabular}}
\end{table}

Three observations follow. First, every pruned graph strongly favors the tokens it retains: excluded tokens receive 19--28\% of the pruned-graph score mass, compared with 72--94\% of the full-graph mass on the same tokens. Second, this bias does not require feedback. Because the independent mask is drawn uniformly, a selection that ignored it would overlap it by $16/256\approx6\%$ in expectation, yet 59\% of the pruned-indep selection stays inside it. Third, the conditioning mask sets how much of the full-graph ranking survives. Starting from the full probe's previous mask, pruned scoring recovers 40\% of the full-graph top-$B$, compared with 32\% from its own previous mask and 22\% from an independent mask, while overlap remains above 80\% for both previous-frame masks. Feedback thus selects which tokens are locked in, and the mask-conditioned graph keeps them.
Re-entry recall is low in all pruned conditions. These are open-loop replays on demonstrations, so they account for the persistence in Table~\ref{tab:graph} but do not directly predict closed-loop success.

\subsection{Joint Student for $\pi_{0.5}$}
\label{app:joint}

The joint student uses the same architecture (140,225 parameters), objective, and optimization as the per-suite students, trained on the union of the four suites' cached labels with their original episode splits (6,838 training and 1,683 validation frames). No labels are recomputed, so it adds no teacher cost. Frames are not reweighted, so Long contributes about 40\% of training data.
The checkpoint with the lowest validation KL on the pooled validation set is selected (epoch 18). Table~\ref{tab:app_joint} compares it with the per-suite students on each suite's validation frames and in closed loop.

\begin{table}[tbp]
    \centering
    \caption{Per-suite and joint students on $\pi_{0.5}$. Validation metrics use each
    suite's validation frames; success uses MMDP at $B=32$ with 500 episodes per suite.
    Per-suite success values are from Table~\ref{tab:main}.}
    \label{tab:app_joint}
    \setlength{\tabcolsep}{4pt}
    \vlaAppendixFit{\linewidth}{%
    \begin{tabular}{llccccc}
    \toprule
     & Student & Spatial & Object & Goal & Long & Avg. \\
    \midrule
    \multirow{2}{*}{Val KL $\downarrow$} & Per-suite & 0.1341 & 0.1281 & 0.1367 & 0.1327 & -- \\
     & Joint & \textbf{0.1335} & \textbf{0.1275} & \textbf{0.1364} & \textbf{0.1324} & -- \\
    \midrule
    \multirow{2}{*}{Recall@32 $\uparrow$} & Per-suite & 0.553 & 0.585 & 0.540 & 0.536 & -- \\
     & Joint & \textbf{0.555} & \textbf{0.586} & \textbf{0.545} & \textbf{0.541} & -- \\
    \midrule
    \multirow{2}{*}{Success (\%)} & Per-suite & \textbf{97.0} & 96.2 & \textbf{93.6} & 76.4 & \textbf{90.80} \\
     & Joint & 96.4 & \textbf{97.0} & 91.0 & \textbf{77.8} & 90.55 \\
    \bottomrule
    \end{tabular}}
\end{table}

The joint student matches or slightly improves offline label fit on every suite. Closed-loop differences range from $-2.6$~pp (Goal) to $+1.4$~pp (Long), with an average of $-0.25$~pp. With 500 episodes, the standard error of a single suite's success rate is about 1.2~pp near 92\%, and per-suite checkpoints vary by 1--3~pp across training seeds (Table~\ref{tab:teacher}), so we do not interpret these differences as systematic. Both students are single checkpoints.

\subsection{Transfer Across Suites and Budgets}
\label{app:transfer}

\begin{table}[tbp]
    \centering
    \caption{Cross-suite transfer on OpenVLA-OFT. A POKE student trained using one suite's
    teacher labels is evaluated on another without retraining. $B=64$ changes
    both suite and budget. Parentheses give differences (pp) from target-suite
    training; italic VP values are target-suite baselines.}
    \label{tab:app_cross_suite}
    \setlength{\tabcolsep}{3pt}
    \vlaAppendixFit{\linewidth}{
    \begin{tabular}{llcccc}
    \toprule
     & & \multicolumn{2}{c}{$B=32$} & \multicolumn{2}{c}{$B=64$} \\
    \cmidrule(lr){3-4}\cmidrule(lr){5-6}
    Evaluate & Train & Top-k & MMDP & Top-k & MMDP \\
    \midrule
    Long    & Spatial & 80.2 ($-2.6$) & 84.4 ($-3.6$) & 89.0 ($-4.4$) & 91.0 ($-1.8$) \\
            & \textit{VP} & \multicolumn{2}{c}{\textit{78.6}} & \multicolumn{2}{c}{\textit{88.6}} \\
    \midrule
    Spatial & Long    & 97.4 ($-0.8$) & 99.4 ($+1.6$) & 97.8 ($-0.4$) & 97.2 ($-1.8$) \\
            & \textit{VP} & \multicolumn{2}{c}{\textit{94.0}} & \multicolumn{2}{c}{\textit{96.0}} \\
    \bottomrule
    \end{tabular}}
\end{table}

Across the four rule/budget combinations, Spatial-to-Long transfer loses 3.1 pp on average relative to target-suite training, whereas the reverse direction loses approximately 0.4 pp. Scene diversity may explain the asymmetry, but we did not separate it from other suite differences.

\section{Autoregressive Policy: OpenVLA}
\label{app:openvla}

We apply POKE to OpenVLA~\cite{kim2024openvla}, retaining $B$ of 256 third-person visual tokens at the input to layer 3. Sensitivities use the continuous relaxation in Appendix~\ref{app:policies} ($A=7$). We evaluate $B=32$.

\begin{table}[tbp]
    \centering
    \caption{LIBERO success rates (\%) for autoregressive OpenVLA at $B=32$ of
    256 visual tokens, with 500 episodes per suite. Parentheses in online-probe
    names indicate the scoring graph. Bold marks the highest non-Vanilla value in
    each column, including ties.}
    \label{tab:app_openvla}
    \label{tab:openvla}
    \setlength{\tabcolsep}{4pt}
    \vlaAppendixFit{\linewidth}{%
    \begin{tabular}{lccccc}
    \toprule
    Method & Spatial & Object & Goal & Long & Avg. \\
    \midrule
    Vanilla                & 84.6 & 84.0 & 74.0 & 51.4 & 73.50 \\
    Random                 & 12.2 & 41.2 & 31.2 & 8.8 & 23.35 \\
    FastV                  & 61.2 & 74.2 & 58.8 & 36.4 & 57.65 \\
    VLA-Cache              & 67.4 & 71.0 & 60.2 & 34.2 & 58.20 \\
    VLA-Pruner             & 75.6 & 76.2 & 61.4 & 42.0 & 63.80 \\
    Probe top-k (full)     & 72.4 & 76.2 & 66.4 & 43.4 & 64.60 \\
    Probe MMDP (full)      & 78.0 & 80.8 & 72.2 & \textbf{53.6} & 71.15 \\
    POKE (top-$k$)          & 72.8 & 81.2 & 70.0 & 44.2 & 67.05 \\
    POKE (MMDP)           & \textbf{78.8} & \textbf{82.2} & \textbf{72.4} & \textbf{53.6} & \textbf{71.75} \\
    \bottomrule
    \end{tabular}}
\end{table}

At $B=32$ (Table~\ref{tab:openvla}), POKE (MMDP) reaches 71.75\% average success, 97.6\% of Vanilla (73.50\%), and exceeds VP (63.80\%) by 7.95~pp, with gains on all four suites (+3.2 to +11.6~pp, largest on Goal and Long). POKE (top-$k$) reaches 67.05\%, so diversity selection adds 4.70~pp, between the gains on OpenVLA-OFT (1.35~pp) and $\pi_{0.5}$ (11.75~pp).
Full-graph probe MMDP reaches 71.15\%, within 0.6~pp of POKE (MMDP), which needs no online backpropagation. On Goal (Table~\ref{tab:openvla_latency}), POKE (MMDP) runs at $1.35\times$, comparable to VP ($1.34\times$), whereas full-graph probe top-$k$ is slower than Vanilla ($0.64\times$).

The squared-sensitivity estimator has lower relative variance at $A=7$ than at $A=56$ (1.33 vs.\ 1.90), although single-probe noise remains substantial (Appendix~\ref{app:probe_var}). For a fixed token pair, Eq.~\eqref{eq:flipprob} depends on the norm ratio $r$ and alignment $\rho$, not directly on $A$. Both MMDP variants exceed Vanilla on Long (53.6 vs.\ 51.4\%); we do not read this as evidence that pruning improves the underlying policy.

\begin{table}[tbp]
    \centering
    \caption{Latency (ms/call) and speedup over Vanilla on OpenVLA Goal ($B=32$),
    using the protocol in Appendix~\ref{app:eval}.}
    \label{tab:app_openvla_latency}
    \label{tab:openvla_latency}
    \setlength{\tabcolsep}{6pt}
    \vlaAppendixFit{\linewidth}{%
    \begin{tabular}{lcc}
    \toprule
    Method & ms & Speedup \\
    \midrule
    Vanilla            & 125.8 & 1.00$\times$ \\
    VLA-Pruner         &  94.2 & 1.34$\times$ \\
    VLA-Cache          & 101.1 & 1.24$\times$ \\
    Probe top-k (full) & 197.8 & 0.64$\times$ \\
    POKE (top-$k$)      & \textbf{88.2} & \textbf{1.43}$\times$ \\
    POKE (MMDP)       &  93.1 & 1.35$\times$ \\
    \bottomrule
    \end{tabular}}
\end{table}

\section{SimplerEnv with \texorpdfstring{$\pi_0$}{pi0}}
\label{app:simpler}

\paragraph{Policy and environment.}
We use open-pi-zero~\cite{ren2024openpizero}, a PyTorch re-implementation of $\pi_0$~\cite{black2024pi0}, with its public Bridge checkpoint (Beta time sampling), trained on BridgeData V2~\cite{walke2023bridgedata}. It takes one $224\times224$ third-person image (256 SigLIP tokens), at most 20 text tokens, and one proprioceptive token, and predicts a chunk of four 7-D actions with ten Euler steps of its action expert against the cached prefix. We evaluate the four WidowX visual-matching tasks of SimplerEnv~\cite{li24simpler} with the authors' protocol: 24 object configurations per task, 10 trials each (240 episodes), all four predicted actions executed, and success read at the time limit. With open-pi-zero's own evaluation code in float32, Vanilla reaches 57.5/85.0/82.9/45.4\% (Carrot, Eggplant, Spoon, Stack; 67.7\% on average) against the reported 55.8/85.4/84.6/47.9\% (68.4\%).

\paragraph{Pruning.}
We delete visual tokens at the input to layer 3 of the 18-layer Gemma backbone, the same cut as for $\pi_{0.5}$, keeping $B=32$ of 256 tokens; retained tokens keep their original position ids, and the action expert reads the reduced cache at every flow step. As on LIBERO, the first three control steps of each episode are unpruned, and the initial flow noise is fixed per (task, episode, control step), so all methods share the noise at matching control steps and outcomes are paired by episode. Our pruned forward pass re-implements the layer loop with the policy's own modules; without pruning it reproduces open-pi-zero's action output exactly under the same noise, and the resulting Vanilla (68.2\%) agrees with the reproduction above within 0.5~pp. FastV reads attention at layer 2; VLA-Pruner follows its $\pi_0$ setting (Appendix~\ref{app:baselines}), with action-to-visual attention averaged over the latter half of the expert layers and flow steps and a window of three control steps ($\gamma=0.8$).

\paragraph{Teacher and student.}
Teacher labels follow the $\pi_{0.5}$ protocol: one probe through all ten flow steps of the full graph, camera-mean baseline, and the final (unclipped) chunk of $A=28$ dimensions in executed units. Labels come from the policy's training data: eight shards spread evenly over the 1{,}024 training shards of BridgeData V2, skipping episodes without instructions, which gives 300 episodes and 2{,}758 frames at a stride of four environment steps (the control interval). This takes 0.82\,s per frame (38 minutes) on one GPU. The student has the architecture of Section~\ref{sec:method_student} with one camera and a 7-D proprioceptive input (140{,}161 parameters) and is trained as in Appendix~\ref{app:student_details} on an 80/20 episode split (2{,}242/516 frames); the selected checkpoint reaches a validation top-32 recall of 0.61 (0.125 by chance). The student is thus trained on real images and deployed on simulated ones without adaptation.

\paragraph{Results.}
Table~\ref{tab:app_simpler_paired} pairs each method with VLA-Pruner by episode. POKE (MMDP) succeeds on 152 episodes where VLA-Pruner fails and fails on 91 where it succeeds (exact McNemar $p=1.1\times10^{-4}$ over 960 episodes), with significant gains on Eggplant and Stack. POKE (top-$k$) ties VLA-Pruner (44.27\% each), so on this flow policy, as on $\pi_{0.5}$, score filtering alone does not separate POKE from the attention baselines and diversity selection over the candidates does. Eggplant, whose sink scene places the target far from the gripper, drops most for every method; Vanilla keeps 84.6\%, POKE (MMDP) 30.8\%, VLA-Pruner 20.4\%, and FastV 7.5\%.

\begin{table}[h]
    \centering
    \caption{\textbf{SimplerEnv paired comparison with VLA-Pruner} ($\pi_0$, $B=32$).
    Entries: episodes where only the method succeeds / only VLA-Pruner succeeds
    (two-sided exact McNemar $p$), 240 episodes per task.}
    \label{tab:app_simpler_paired}
    \setlength{\tabcolsep}{3pt}
    \resizebox{\linewidth}{!}{%
    \begin{tabular}{lccccc}
    \toprule
    Method & Carrot & Eggplant & Spoon & Stack & Pooled \\
    \midrule
    FastV          & 19/24 (0.54) & 9/40 ($9.3{\times}10^{-6}$) & 21/37 (0.048) & 40/31 (0.34) & 89/132 (0.0046) \\
    POKE (top-$k$) & 22/24 (0.88) & 37/38 (1.00) & 30/45 (0.11) & 39/21 (0.027) & 128/128 (1.00) \\
    POKE (MMDP)    & 19/21 (0.87) & 52/27 (0.0066) & 35/24 (0.19) & 46/19 (0.0011) & 152/91 ($1.1{\times}10^{-4}$) \\
    \bottomrule
    \end{tabular}}
\end{table}

\section{LIBERO-Plus Evaluation}
\label{app:liberoplus}

\paragraph{Protocol.}
LIBERO-Plus~\cite{fei2025liberoplus} provides 2,402, 2,518, 2,591, and 2,519 perturbed variants of the Spatial, Object, Goal, and Long tasks. Within each suite we group tasks by perturbation category and difficulty level, fix a random order within each group, and take tasks in proportion to group size, giving 600, 630, 648, and 630 tasks (25\%). Every method and both policies evaluate exactly these tasks with the initial state of the official evaluation and one episode per task. Episode limits, the settle steps, action chunking, the cut layer, the per-camera budget, and the three unpruned warmup control steps are those of the main experiments. For $\pi_{0.5}$, initial noise is seeded by (suite, task, episode, control step) as in Appendix~\ref{app:eval}. The policies, POKE students, and baselines are those of Table~\ref{tab:main}; no component is retrained or re-tuned on perturbed data. Retention is a method's success divided by Vanilla success on the same tasks.

Table~\ref{tab:libero_plus_suite} reports per-suite success for the same runs. On OpenVLA-OFT, both POKE variants exceed VP on every suite and MAP$^\dagger$ on every suite but Goal, where all three are within 0.5~pp.

\begin{table}[t]
\centering
\caption{LIBERO-Plus success (\%) by suite at $B=32$. Avg.: unweighted mean over suites; Ret.: Avg.\ relative to Vanilla. The Total column of Table~\ref{tab:liberoplus} pools tasks across suites and can differ slightly.}
\label{tab:libero_plus_suite}
\resizebox{\linewidth}{!}{%
\begin{tabular}{clcccccc}
\toprule
 & Method & Spatial & Object & Goal & Long & Avg.($\uparrow$) & Ret.($\uparrow$) \\
\midrule
\multirow{6}{*}{\rotatebox[origin=c]{90}{\footnotesize OpenVLA-OFT}}
 & Vanilla        & 82.7 & 83.8 & 64.0 & 67.8 & 74.6 & 1.00 \\
 & FastV          & 77.0 & 33.2 & 44.4 & 36.0 & 47.7 & 0.64 \\
 & DivPrune       & 53.3 & 27.8 & 42.6 & 19.4 & 35.8 & 0.48 \\
 & VLA-Pruner     & 71.7 & 63.3 & 50.6 & 50.5 & 59.0 & 0.79 \\
 & MAP$^\dagger$  & 71.8 & 63.2 & 59.3 & 50.8 & 61.3 & 0.82 \\
 & POKE (top-$k$) & \textbf{81.3} & \textbf{75.7} & \textbf{59.6} & \textbf{58.6} & \textbf{68.8} & \textbf{0.92} \\
 & POKE (MMDP)    & 79.7 & 73.3 & 59.1 & 57.3 & 67.4 & 0.90 \\
\midrule
\multirow{6}{*}{\rotatebox[origin=c]{90}{$\pi_{0.5}$}}
 & Vanilla        & 91.0 & 87.8 & 78.2 & 77.9 & 83.7 & 1.00 \\
 & FastV          & 45.7 & 42.5 & 38.0 & 17.1 & 35.8 & 0.43 \\
 & DivPrune       & 54.7 & 51.3 & 44.3 & 24.1 & 43.6 & 0.52 \\
 & VLA-Pruner     & 44.5 & 51.1 & 43.5 & 24.9 & 41.0 & 0.49 \\
 & MAP$^\dagger$  & 42.5 & 49.8 & 43.4 & 21.6 & 39.3 & 0.47 \\
 & POKE (top-$k$) & 68.3 & 60.8 & 50.5 & 31.3 & 52.7 & 0.63 \\
 & POKE (MMDP)    & \textbf{83.3} & \textbf{72.5} & \textbf{64.0} & \textbf{53.7} & \textbf{68.4} & \textbf{0.82} \\
\bottomrule
\end{tabular}}
\end{table}

\section{Detailed Runtime Results}
\label{app:latency}

\begin{table*}[tbp]
    \centering
    \caption{OpenVLA-OFT latency (ms/call) and paired speedups over Vanilla,
    using Appendix~\ref{app:eval}. Only Goal is timed at $B=128$.
    The final column reports speedups only.}
    \label{tab:app_latency}
    \vlaAppendixFit{\textwidth}{
    \begin{tabular}{llcccccc}
    \toprule
    Suite & $B$ & Vanilla & VLA-Pruner & FastV & POKE (top-$k$) & POKE (MMDP) & Probe (pruned) \\
    \midrule
    Spatial & 32 & 129.9 & 70.7 / 1.84$\times$ & 68.6 / 1.90$\times$ & 68.1 / 1.91$\times$ & 69.7 / 1.87$\times$ & 1.16$\times$ \\
    Object  & 32 & 129.0 & 70.7 / 1.82$\times$ & 68.3 / 1.88$\times$ & 68.0 / 1.90$\times$ & 69.6 / 1.85$\times$ & 1.05$\times$ \\
    Goal    & 32 & 130.1 & 71.5 / 1.82$\times$ & 69.2 / 1.88$\times$ & 68.8 / 1.89$\times$ & 70.4 / 1.85$\times$ & 1.05$\times$ \\
    Long    & 32 & 130.7 & 72.0 / 1.82$\times$ & 69.2 / 1.89$\times$ & 69.2 / 1.89$\times$ & 70.9 / 1.85$\times$ & 1.06$\times$ \\
    \midrule
    Spatial & 64 & 130.7 & 79.2 / 1.65$\times$ & 75.9 / 1.72$\times$ & 75.5 / 1.73$\times$ & 78.1 / 1.68$\times$ & 0.97$\times$ \\
    Object  & 64 & 130.4 & 78.6 / 1.66$\times$ & 75.5 / 1.72$\times$ & 75.0 / 1.73$\times$ & 77.6 / 1.68$\times$ & 0.97$\times$ \\
    Goal    & 64 & 130.3 & 78.6 / 1.66$\times$ & 75.5 / 1.73$\times$ & 74.9 / 1.74$\times$ & 77.8 / 1.68$\times$ & 0.97$\times$ \\
    Long    & 64 & 130.4 & 78.1 / 1.66$\times$ & 75.2 / 1.73$\times$ & 74.8 / 1.74$\times$ & 77.3 / 1.68$\times$ & 0.97$\times$ \\
    \midrule
    Goal    & 128 & 129.9 & 97.5 / 1.33$\times$ & 92.9 / 1.40$\times$ & 91.9 / 1.42$\times$ & 96.7 / 1.35$\times$ & 0.84$\times$ \\
    \bottomrule
    \end{tabular}
    }
\end{table*}

POKE (top-$k$) is faster than POKE (MMDP) in every reported suite/budget condition, consistent with MMDP's additional candidate-distance computation. The difference is approximately 1.6\,ms at $B=32$, 2.7\,ms at $B=64$, and 4.8\,ms at $B=128$. POKE (MMDP) beats VP in 83\% of timing repetitions at $B=128$, versus 93--98\% in the other settings; top-k beats VP in at least 98\% throughout.

Pruned probes exhibit greater timing variation: $B=32$ speedups span $1.04$--$1.16\times$ across the recorded runs, with Spatial near the upper end. At $B=64,128$, the reported ranges narrow to less than $0.01\times$, but speedups fall below one. Other methods vary within approximately $\pm0.02\times$ across the four suites.

\begin{table}[tbp]
    \centering
    \caption{Success rates (\%) when reusing masks for $R$ control steps:
    OpenVLA, $B=32$, online full-graph probe MMDP. Scoring once every $R$ control
    steps reduces its amortized frequency to $1/R$, while masks become stale.}
    \label{tab:app_k}
    \setlength{\tabcolsep}{6pt}
    \vlaAppendixFit{\linewidth}{%
    \begin{tabular}{lcccc}
    \toprule
    Suite & $R=1$ & $R=2$ & $R=4$ & $R=8$ \\
    \midrule
    Spatial & 78.0 & 77.0 & 77.0 & 71.4 \\
    Object  & 80.8 & 81.8 & 80.6 & 77.0 \\
    Goal    & 72.2 & 72.6 & 70.8 & 67.4 \\
    \bottomrule
    \end{tabular}}
\end{table}

Success changes little at $R=2$, but drops on all three suites at $R=8$ ($-6.6$, $-3.8$, and $-4.8$ pp relative to $R=1$).
Mask reuse thus trades scoring cost for stale selections, whereas the POKE student scores every observation.

\subsection{$\pi_{0.5}$ Timing}
\label{app:latency_pi05}

The eager PyTorch path spends most of its measured time in the ten-step action-expert loop (Table~\ref{tab:app_lat_stage_pi05}). We use CUDA graphs for the image encoder, VLM prefix, and denoising loop to reduce launch overhead. The prefix is split around selection into layers $0$--$k-1$ and $k$--17. FastV selection, VP's combine-then-filter operation, student scoring, and MMDP run outside the graphs and remain within the timing boundary. For the tested identical inputs, graph-path actions and VP action attention match the eager outputs with maximum absolute error zero, including FastV/VP selection. Online-probe backward passes are not captured. All methods, including Vanilla, online probes, and POKE, share one re-implemented path that omits the masked camera slot and prompt padding, so ratios are unaffected. This Vanilla is faster than openpi's reference path (334\,ms per call), so using it as the denominator is conservative. Timings use an RTX A6000 and five processes with 50 repetitions each. Online-probe backward passes are not captured, so all probe timings on $\pi_{0.5}$ use the eager path and are compared with eager Vanilla (Table~\ref{tab:app_lat_eager_pi05}).

\begin{table}[tbp]
    \centering
    \caption{Per-stage $\pi_{0.5}$ latency (ms, $B=32$; median of 20 repetitions per entry).
    Image encoding, the VLM prefix, the 10-step denoising loop, and the full sampling call are each
    timed with CUDA events and device synchronization around the call, on a synthetic observation
    with a fixed keep set (no token selection). All entries come from one session; in every row the
    stage sum matches the full call within 0.5\,ms. The full call excludes input/output transforms,
    token selection, and the device-to-host copy, which the end-to-end timings in
    Table~\ref{tab:latency} include (83.2 vs.\ 81.4\,ms for Vanilla under CUDA graphs in the same
    session). Prefix GFLOPs count the VLM backbone only.}
    \label{tab:app_lat_stage_pi05}
    \vlaAppendixFit{\linewidth}{
    \begin{tabular}{ll cccc c}
    \toprule
    Execution & Path & Image & Prefix & Denoise & Full call & Prefix GFLOPs \\
    \midrule
    Eager      & Full   & 27.4 & 29.7 & 232.0 & 289.0 & 2{,}101 \\
               & $B=32$ & 27.4 & 18.0 & 232.4 & 278.1 & 482 \\
    \midrule
    CUDA graph & Full   &  9.7 & 28.5 &  42.9 &  81.4 & 2{,}101 \\
               & $B=32$ &  9.7 & 11.6 &  41.5 &  62.3 & 482 \\
    \bottomrule
    \end{tabular}
    }
\end{table}

\begin{table}[tbp]
    \centering
    \caption{$\pi_{0.5}$ latency (ms) and speedups in eager mode, where every method,
    including online-probe backward passes, runs on the same execution path. Speedups
    are medians of paired ratios over five processes (ranges in brackets); Vanilla
    omits masked camera/prompt padding as in Table~\ref{tab:latency}.}
    \label{tab:app_lat_eager_pi05}
    \setlength{\tabcolsep}{3pt}
    \vlaAppendixFit{\linewidth}{
    \begin{tabular}{l cc cc cc}
    \toprule
     & \multicolumn{2}{c}{$B=32$} & \multicolumn{2}{c}{$B=64$} & \multicolumn{2}{c}{$B=128$} \\
    \cmidrule(lr){2-3}\cmidrule(lr){4-5}\cmidrule(lr){6-7}
    Method & ms & Speedup & ms & Speedup & ms & Speedup \\
    \midrule
    Vanilla              & 300.0 & 1.00$\times$ & 306.0 & 1.00$\times$ & 299.9 & 1.00$\times$ \\
    FastV                & 290.3 & 1.03$\times$ [1.03--1.04] & 296.7 & 1.03$\times$ [1.03--1.04] & 290.5 & 1.03$\times$ [1.03--1.03] \\
    VLA-Pruner           & 296.9 & 1.01$\times$ [1.01--1.01] & 304.6 & 1.00$\times$ [1.00--1.01] & 301.0 & 1.00$\times$ [0.99--1.00] \\
    POKE (top-$k$)       & 290.7 & 1.03$\times$ [1.03--1.03] & 297.2 & 1.03$\times$ [1.03--1.04] & 290.9 & 1.03$\times$ [1.03--1.03] \\
    POKE (MMDP)          & 292.5 & 1.03$\times$ [1.02--1.03] & 300.4 & 1.02$\times$ [1.02--1.02] & 296.6 & 1.01$\times$ [1.01--1.01] \\
    Probe top-k (pruned) & 636.3 & 0.48$\times$ [0.46--0.49] & 646.3 & 0.47$\times$ [0.46--0.49] & 638.7 & 0.48$\times$ [0.46--0.49] \\
    Probe MMDP (pruned)  & 638.1 & 0.48$\times$ [0.46--0.49] & 650.7 & 0.47$\times$ [0.46--0.49] & 645.6 & 0.47$\times$ [0.46--0.48] \\
    Probe top-k (full)   & 925.9 & 0.32$\times$ [0.32--0.33] & 951.0 & 0.32$\times$ [0.32--0.33] & 939.4 & 0.32$\times$ [0.32--0.33] \\
    \bottomrule
    \end{tabular}}
\end{table}

In eager mode the denoising loop dominates (Table~\ref{tab:app_lat_stage_pi05}), so pruning yields at most $1.03\times$, whereas the full-graph probe runs at $0.32\times$; CUDA graphs remove launch overhead and expose the prefix savings.

\begin{table*}[tbp]
    \centering
    \caption{Latency (ms/call) and speedups at $B=32$. Speedups are medians of paired
    ratios with randomized method order; brackets give ranges over five independent
    processes. For $\pi_{0.5}$, the image encoder, prefix, and denoising loop use
    CUDA graphs; selection and probe backward passes run outside them.
    Vanilla uses the same path with masked camera/prompt padding omitted.
    See Appendix~\ref{app:eval}.}
    \label{tab:app_latency_main}
    \label{tab:latency}
    \vlaAppendixFit{\textwidth}{
    \begin{tabular}{lcccc}
    \toprule
     & \multicolumn{2}{c}{OpenVLA-OFT} & \multicolumn{2}{c}{$\pi_{0.5}$} \\
    \cmidrule(lr){2-3}\cmidrule(lr){4-5}
    Method & ms & Speedup & ms & Speedup \\
    \midrule
    Vanilla              & 130.1 & 1.00$\times$ & 84.7 & 1.00$\times$ \\
    FastV                &  69.2 & 1.88$\times$ [1.86--1.88] & \textbf{65.6} & \textbf{1.29}$\times$ [1.29--1.29] \\
    VLA-Pruner           &  71.5 & 1.82$\times$ [1.80--1.82] & 68.5 & 1.24$\times$ [1.24--1.24] \\
    VLA-Cache            &  88.6 & 1.47$\times$ [1.47--1.48] & -- & -- \\
    Probe top-k (pruned) & 123.0 & 1.05$\times$ [1.04--1.15] & 630.2 & 0.13$\times$ [0.13--0.14] \\
    Probe MMDP (pruned)  & 124.8 & 1.04$\times$ [1.03--1.14] & 632.4 & 0.13$\times$ [0.13--0.14] \\
    Probe top-k (full)   & 321.0 & 0.41$\times$ [0.41--0.41] & 700.4 & 0.12$\times$ [0.12--0.13] \\
    POKE (top-$k$)        & \textbf{68.8} & \textbf{1.89}$\times$ [1.88--1.89] & 65.8 & \textbf{1.29}$\times$ [1.29--1.29] \\
    POKE (MMDP)         &  70.4 & 1.85$\times$ [1.84--1.85] & 67.4 & 1.26$\times$ [1.26--1.26] \\
    \bottomrule
    \end{tabular}
    }
\end{table*}

\begin{table}[tbp]
    \centering
    \caption{Budget-dependent $\pi_{0.5}$ latency (ms) and speedups with CUDA graphs.
    FLOPs are percentages of the unpruned VLM backbone; values for $B=64,128$
    are estimates from layer-wise token counts.}
    \label{tab:app_lat_budget_pi05}
    \vlaAppendixFit{\linewidth}{
    \begin{tabular}{l cc cc cc}
    \toprule
     & \multicolumn{2}{c}{$B=32$} & \multicolumn{2}{c}{$B=64$} & \multicolumn{2}{c}{$B=128$} \\
    \cmidrule(lr){2-3}\cmidrule(lr){4-5}\cmidrule(lr){6-7}
    Method & ms & Speedup & ms & Speedup & ms & Speedup \\
    \midrule
    Vanilla              & 84.7 & 1.00$\times$ & 84.6 & 1.00$\times$ & 84.8 & 1.00$\times$ \\
    FastV                & 65.6 & 1.29$\times$ & 66.6 & 1.27$\times$ & 69.5 & 1.22$\times$ \\
    VLA-Pruner           & 68.5 & 1.24$\times$ & 70.5 & 1.20$\times$ & 75.4 & 1.12$\times$ \\
    Probe top-k (pruned) & 630.2 & 0.13$\times$ & 625.0 & 0.14$\times$ & 629.9 & 0.13$\times$ \\
    Probe MMDP (pruned)  & 632.4 & 0.13$\times$ & 628.8 & 0.13$\times$ & 635.7 & 0.13$\times$ \\
    Probe top-k (full)   & 700.4 & 0.12$\times$ & 697.2 & 0.12$\times$ & 704.4 & 0.12$\times$ \\
    POKE (top-$k$)        & 65.8 & 1.29$\times$ & 66.8 & 1.27$\times$ & 69.6 & 1.22$\times$ \\
    POKE (MMDP)         & 67.4 & 1.26$\times$ & 69.5 & 1.22$\times$ & 74.7 & 1.14$\times$ \\
    \midrule
    Backbone FLOP        & \multicolumn{2}{c}{22.9\%} & \multicolumn{2}{c}{28.1\%} & \multicolumn{2}{c}{38.3\%} \\
    \bottomrule
    \end{tabular}
    }
\end{table}

Under CUDA graphs, pruning to $B=32$ makes the prefix $2.46\times$ faster (28.5 to 11.6\,ms), consistent with the $4.4\times$ reduction in prefix FLOPs, but leaves the denoising loop nearly unchanged (42.9 to 41.5\,ms). The loop runs the action expert ten times over a short action sequence, so shortening the cached prefix barely changes its cost, and it accounts for two thirds of the pruned call. The full call therefore speeds up by $1.31\times$, and the end-to-end $1.26\times$ in Table~\ref{tab:latency} additionally includes transforms and token selection. The same structure explains the small latency change between $B=128$ and $B=32$ (Table~\ref{tab:app_lat_budget_pi05}).

VLA-Pruner reports $1.95\times$ on $\pi_0$ at 12.5\% retention (RTX~4090; Vanilla 104.5~ms). That result uses a different policy, implementation, and hardware, and we did not reproduce its pipeline. In ours, the 35\% prefix share bounds any token pruner at $1.54\times$.

VP costs 2.9--5.9\,ms more than FastV, while POKE (MMDP) costs 1.6--5.1\,ms more than POKE (top-$k$) across budgets. The latter differs from FastV by at most 0.2\,ms. These are differences between whole pipelines, not direct measurements of the student scorer's cost. Selection overhead grows with budget in these measurements, but we did not characterize how it scales.

\section{Statistical Analyses}
\label{app:stats}

Episode-paired tests match initial states. Bootstrap intervals resample paired episodes within fixed tasks. Sign tests use tasks as units. We do not treat similar point estimates as evidence of equivalence or of improvement unless a test supports it.

\begin{table*}[tbp]
    \centering
    \caption{Paired-episode analyses of the scoring graph on OpenVLA-OFT Goal
    ($B=32$). A only counts episodes where A succeeds and B fails. Exact $p$
    values use a two-sided binomial test on discordant pairs. Holm correction
    treats the three comparisons as one family.}
    \label{tab:app_paired}
    \vlaAppendixFit{\textwidth}{
    \begin{tabular}{lcccccc}
    \toprule
    Comparison (A $-$ B) & A only & B only & Difference (pp) & Paired 95\% CI & Exact $p$ & Holm $p$ \\
    \midrule
    Full $-$ Pruned, top-k             & 40 & 9  & $+6.2$ & $[3.6, 8.8]$ & $9.3\times10^{-6}$ & $2.8\times10^{-5}$ \\
    Full top-k $-$ Pruned MMDP         & 35 & 12 & $+4.6$ & $[2.0, 7.4]$ & $1.1\times10^{-3}$ & $2.2\times10^{-3}$ \\
    Pruned MMDP $-$ Pruned top-k       & 34 & 26 & $+1.6$ & $[-1.4, 4.6]$ & 0.366 & 0.366 \\
    \bottomrule
    \end{tabular}
    }
\end{table*}

The Goal graph effect remains significant after correction, and full-graph top-k also exceeds pruned-graph MMDP. The same-session selection-rule comparison under pruned scoring is not significant.

For Long's $+22.2$ pp graph effect, nine of ten tasks improve (two-sided sign test $p=0.0215$). An unpaired approximation gives $z=7.95$, but it is not an episode-paired analysis, so the main text reports effect size and task-wise direction instead.

\section{Surrogate Fidelity and Offline Mask Quality}
\label{app:surrogate}

\paragraph{Three evaluation sets.}
All analyses in this section use OFT, $B=32$, and deletion at layer 4.
\begin{itemize}
    \item \textbf{Set 1: mask quality.} We apply alternative masks to 120 frames
    from 43 Goal validation episodes, sampled at frame indices 3/8/13/18/23
    after warmup. Deviations are normalized by the policy's dimension-wise
    output ranges. An initial filename-ordered sample mostly contained training
    episodes; its POKE results were discarded. All reported Set 1 results
    were rerun on validation episodes.
    \item \textbf{Set 2: replacement and probe analysis.} On 100 validation frames
    from Spatial, we compute and store exact $u_i=J_i\delta_i$ at the
    configured model precision. We resample probes across ten seeds without
    recomputing the Jacobian. Replacement/deletion comparisons use seven masks
    per frame: exact top-k, three probe top-k masks, and three random masks
    (700 frame--mask pairs).
    \item \textbf{Set 3: shared-observation comparisons.} Vanilla closed-loop
    rollouts on Long use ten tasks and two prespecified initial states per task
    (indices 0 and 1). After warmup, FastV, VP, both POKE variants, and three random
    masks are evaluated on each of the 625 shared observations, one per control step.
    These evaluation forwards do not control the environment. VP updates its
    temporal signal along the entire Vanilla trajectory, including warmup.
    Deviation is the chunk-wide L2 norm after normalization by dimension-wise
    q01--q99 ranges, divided by the mean deviation of three random masks on the
    same frame.
\end{itemize}
Frames and masks are repeated measurements. We report distributions for Set 2 without tests that assume independent frame--mask pairs. For Set 3, we aggregate paired frame-wise deviation differences within each task by the median and apply sign tests at the task level.

\paragraph{Single-probe variance.}
\label{app:probe_var}
For a uniform spherical $v\in\mathbb R^A$ and unit $\hat u$,
$\mathbb E[(v^\top\hat u)^2]=1/A$ and
$\mathbb E[(v^\top\hat u)^4]=3/[A(A+2)]$.
Thus $\widehat s_i^2=A(v^\top u_i)^2$ is unbiased for $s_i^2$ and, for $s_i>0$,
\begin{equation}
    \begin{aligned}
     \frac{\operatorname{Var}[\widehat s_i^2]}{(\mathbb E[\widehat s_i^2])^2}
     &=\frac{3A}{A+2}-1\\
     &=\frac{2(A-1)}{A+2}.
    \end{aligned}
\end{equation}
This ratio increases with $A$ toward two (1.33 at $A=7$, 1.90 at $A=56$, and 1.92 at $A=70$). Averaging $P$ independent squared-score probes divides the relative variance by $P$; the relative standard deviation decreases as $1/\sqrt P$.
The ranking argument in \S\ref{sec:method_probe} instead relies on shared multiplicative factors.

\paragraph{Deriving the reversal probability.}
With a shared probe, $\widehat s_i=\sqrt A\,s_i\kappa_i$ with $\kappa_i=|v^\top\hat u_i|$. For $s_i>s_j>0$, the probe reverses the pair exactly when $\kappa_j/\kappa_i>r$, apart from the zero-probability event $\kappa_i=0$.
Fix nonzero effects with $s_i>s_j$, $r=s_i/s_j>1$, $\rho=\hat u_i^\top\hat u_j$, and $A\ge2$. For $|\rho|<1$, write $v=z/\|z\|$, $z\sim\mathcal N(0,I_A)$. Normalization cancels from the ratio of projections. Then $X=z^\top\hat u_i$ and $Y=z^\top\hat u_j$ are standard normal with correlation $\rho$. With independent standard-normal $W$, $Y=\rho X+\sqrt{1-\rho^2}W$. Since $W/X$ is standard Cauchy, $Y/X$ is Cauchy with location $\rho$ and scale $\sigma=\sqrt{1-\rho^2}$. Its two tails give
\begin{equation}
    \begin{aligned}
     \Pr[|Y/X|>r]
     &=1-\frac1\pi\biggl[\arctan\frac{r-\rho}{\sigma}\\
     &\hspace{4.2em}+\arctan\frac{r+\rho}{\sigma}\biggr].
    \end{aligned}
\end{equation}
For $r>1$, the two positive arctangent arguments have product greater than one. The arctangent addition identity therefore yields
\begin{equation}
 \Pr_v[\widehat s_j>\widehat s_i]
 =\frac1\pi\arctan\!\left(\frac{2r\sqrt{1-\rho^2}}{r^2-1}\right).
 \label{eq:flipprob}
\end{equation}
This expression is independent of $A$ conditional on $(r,\rho)$. At $\rho=0$ it is exactly $\frac2\pi\arctan(1/r)$; at $|\rho|=1$, the common absolute projection factor gives zero reversal probability for a strict norm ordering. The reported Monte Carlo check uses $2\times10^5$ probes per combination of $A\in\{7,56,70\}$, $r\in\{1.5,3,10\}$, and $\rho\in\{0,0.5,-0.9\}$, with absolute errors below 0.003.

If effect directions are themselves sampled independently and uniformly, one additionally averages over their alignment distribution. Here $\mathbb E[\rho^2]=1/A$, and each conditional reversal probability is at most its orthogonal value. The reported direction-averaged values at $r=3$ are 0.192 for $A=7$ and 0.203 for $A=56$, versus 0.205 for $\rho=0$. This dimension-dependent mixture is distinct from the fixed-pair result; actual policy alignment is measured in Table~\ref{tab:app_align}.

\paragraph{Noise dependence of flow-policy labels.}
\label{app:eps}
Each $\pi_{0.5}$ teacher label uses one initial noise $\varepsilon$ and one probe $v$. On 200 Long validation frames, we resample (i) only $\varepsilon$ with $v$ fixed to its teacher value, (ii) only $v$ with $\varepsilon$ fixed, or (iii) both, using eight full-graph samples per condition.
Pairwise metrics average comparisons between sample pairs from the same frame. Against-mean metrics compare each sample to the mean of its eight score vectors, including that sample. Spearman is computed within cameras; recall uses the top-16 per camera, for 32 tokens total.
Recomputing cached labels with identical $(\varepsilon,v)$ gives maximum relative difference zero in the reported check.

\begin{table}[tbp]
    \centering
    \caption{Resampling stability of $\pi_{0.5}$ teacher labels: 200 Long validation frames and eight samples per condition. Values are means across frames, with between-frame SDs in parentheses. Recall uses 16 tokens per camera (32 in total).}
    \label{tab:app_eps}
    \setlength{\tabcolsep}{4pt}
    \vlaAppendixFit{\linewidth}{%
    \begin{tabular}{lcccc}
    \toprule
     & \multicolumn{2}{c}{Pairwise} & \multicolumn{2}{c}{Against mean} \\
    \cmidrule(lr){2-3}\cmidrule(lr){4-5}
    Resampled & Spearman & Recall@32 & Spearman & Recall@32 \\
    \midrule
    $\varepsilon$ only         & \textbf{0.930} (0.030) & \textbf{0.854} (0.063) & \textbf{0.957} (0.022) & \textbf{0.882} (0.071) \\
    Probe $v$ only             & 0.768 (0.031) & 0.566 (0.058) & 0.864 (0.019) & 0.659 (0.056) \\
    $\varepsilon$ and $v$    & 0.763 (0.029) & 0.560 (0.056) & 0.860 (0.019) & 0.652 (0.060) \\
    \bottomrule
    \end{tabular}}
\end{table}

Rankings are more stable under noise-only resampling than under probe-only resampling, while resampling both gives results close to probe-only. This compares ranking stability only; it is not a variance decomposition and does not show that noise effects are negligible. The student, which does not receive $\varepsilon$, learns an observation-conditioned approximation, and the mixed training objective need not average out its noise dependence exactly.
The analysis covers only Long and does not test whether single-noise and noise-averaged labels lead to different closed-loop success. Against-mean metrics use a reference that includes the sample itself, not an independent estimate of the expected score.

\paragraph{Replacement versus deletion.}
Set 2 compares projector-output replacement (\texttt{replace}) and deletion at the cut (\texttt{delete\_k}) for identical retained sets.

\begin{table}[tbp]
    \centering
    \caption{Replacement versus deletion deviations (Set 2).
    The ratio is $D_{\mathrm{repl}}/D_{\mathrm{delete}}$.
    Within-frame Spearman compares the seven masks evaluated at one observation.}
    \label{tab:app_ratio}
    \setlength{\tabcolsep}{5pt}
    \vlaAppendixFit{\linewidth}{%
    \begin{tabular}{lccc}
    \toprule
    Mask & $n$ & Median ratio & IQR \\
    \midrule
    Overall                       & 700 & 1.076 & [0.996, 1.200] \\
    Exact top-k                & 100 & 1.117 & [0.986, 1.333] \\
    Probe top-k (seed 0/1/2)   & 300 & 1.115--1.139 & -- \\
    Random (seed 0/1/2)        & 300 & 1.034--1.040 & -- \\
    \midrule
    \multicolumn{4}{l}{Within-frame Spearman($D_{\mathrm{repl}}$, $D_{\mathrm{delete}}$): median 0.964, Mean 0.933} \\
    \bottomrule
    \end{tabular}}
\end{table}

Replacement and deletion have strongly associated within-frame damage rankings, with a median deviation ratio of 1.076 (about 8\% higher for replacement). The ratio is somewhat larger for top-k masks. The high correlation supports the surrogate empirically, although the two can still select different tokens. The pooled correlation across 700 frame--mask pairs is 0.965; because it mixes frames and mask types, we report within-frame correlations in the main text. In Set 1, cut-layer deletion produces 17--22\% less deviation than projector deletion on the same retained sets ($D_{\mathrm{delete\_k}}/D_{\mathrm{delete\_0}}=0.78$--$0.83$), supporting later deletion for this one-step fidelity measure.

\paragraph{Single-probe ranking.}\label{app:ranking}
Set 2 compares exact sensitivities with $P=1$ estimates.

\begin{table*}[tbp]
    \centering
    \caption{Single-probe ranking on 100 Set 2 frames. The first four rows report mean$\pm$SD (sample SD) across ten probe seeds (0--9), each summarized by its frame mean. Recall and retained energy compare the probe and exact masks, both formed from the top 16 tokens per camera (32 in total); retained energy is the ratio of the exact $\sum s_i^2$ over the two masks. Energy@32 is the fraction of a frame's total exact $\sum_i s_i^2$ over all 512 tokens captured by the exact top-32 tokens, selected globally or as the top 16 per camera.}
    \label{tab:app_probe}
    \setlength{\tabcolsep}{5pt}
    \vlaAppendixFit{\textwidth}{%
    \begin{tabular}{lc}
    \toprule
    Metric & Value \\
    \midrule
    Spearman($\widehat s^{P=1}$, $s$), All tokens        & $0.835\pm0.006$ \\
    Spearman($\widehat s^{P=1}$, $s$), Within-camera mean  & $0.797\pm0.007$ \\
    Top-32 recall                                        & $0.667\pm0.014$ \\
    Retained energy (vs. exact)                               & $0.834\pm0.012$ \\
    \midrule
    Energy@32, Global top-32 (median, IQR)                  & 0.858 [0.827, 0.898] \\
    Energy@32, Top-16 per camera (median, IQR)              & 0.836 [0.809, 0.875] \\
    \bottomrule
    \end{tabular}}
\end{table*}

\begin{table}[tbp]
    \centering
    \caption{Alignment of action-effect directions (Set 2). Pairwise cosine statistics are averaged within each frame and then across frames. The isotropic reference uses independent uniform unit vectors in $\mathbb R^{56}$.}
    \label{tab:app_align}
    \setlength{\tabcolsep}{5pt}
    \vlaAppendixFit{\linewidth}{%
    \begin{tabular}{lccc}
    \toprule
     & signed $\cos$ & $|\cos|$ & $\cos^2$ \\
    \midrule
    All token pairs      & 0.057 & 0.324 & 0.155 \\
    Top-32 pairs    & 0.051 & 0.386 & 0.211 \\
    Isotropic($A=56$) & 0     & 0.107 & 0.018 \\
    \bottomrule
    \end{tabular}}
\end{table}

Within-camera rank correlation is approximately 0.80, with small variation across the ten sampled probe seeds. Retained sensitivity energy is 83\%, compared with 0.67 set recall, consistent with unequal score contributions among tokens.

Absolute-cosine and squared-cosine means exceed isotropic references by approximately three and nine times overall, and 3.6 and twelve times among top-32 tokens, while signed cosines are close to zero. Effect directions are therefore more aligned up to sign than isotropic vectors, without sharing a single common axis; absolute projections depend only on this unsigned alignment. Neither the alignment nor the energy concentration implies large norm gaps at the selection boundary or recovery of the whole set, and we do not separate their contributions here.

\paragraph{What preserves single-probe rankings?}
By Eq.~\eqref{eq:flipprob}, the pairwise reversal probability decreases with both the sensitivity ratio and the alignment of action effects. At $r=3$, substituting the measured mean $|\cos|$ (0.32--0.39) for $|\rho|$ lowers this probability by only about 4--6\% relative to orthogonal effects. In practice, rankings are preserved mainly because sensitivity energy is concentrated in few tokens (about 84\% in the top 16 per camera). This view predicts weaker discrimination when a larger budget moves the selection boundary into regions where energy is spread more evenly. The small differences between methods at $B=128$ are consistent with this, although we do not test the link directly.

\paragraph{Mask quality.}
Set 1 compares action deviations against random masks with the same budget and camera quotas.

\begin{table}[tbp]
    \centering
    \caption{Action deviations relative to random masks with the same $B=32$ and camera quotas (Set 1, \texttt{delete\_k}; lower is better). \emph{Dev./random}: on each frame, every mask's deviation is divided by the mean deviation of the $n=32$ random masks on that frame, and the table reports the median of this ratio over the 120 frames; all masks on a frame share the denominator, so within-frame orderings and the paired tests in Table~\ref{tab:app_wilcoxon} do not depend on it. \emph{Below random}: the fraction of the $120\times32$ frame--random-mask pairs in which the mask's deviation is strictly lower (ties count against the mask). The Random row is the reference by definition (1.000 and 0.500), not a measured value.}
    \label{tab:app_surrogate_full}
    \label{tab:surrogate}
    \setlength{\tabcolsep}{6pt}
    \vlaAppendixFit{\linewidth}{%
    \begin{tabular}{lcc}
    \toprule
    Mask & Dev./random & Below random \\
    \midrule
    Exact top-k (56 backward passes)        & \textbf{0.250} & 0.988 \\
    POKE (top-$k$) (0 backward passes)       & 0.254 & 0.983 \\
    Probe($P{=}3$) top-k               & 0.267 & 0.988 \\
    Probe($P{=}1$) top-k (teacher)     & 0.285 & 0.986 \\
    POKE (MMDP)                       & 0.343 & 0.977 \\
    Exact MMDP                         & 0.364 & 0.977 \\
    Probe($P{=}1$) MMDP                & 0.402 & 0.973 \\
    DivPrune                           & 0.758 & 0.734 \\
    Random                             & 1.000 & 0.500 \\
    Probe($P{=}1$) bottom-k            & 1.599 & 0.044 \\
    Exact bottom-k                     & 1.844 & 0.014 \\
    \bottomrule
    \end{tabular}}
\end{table}

Median range-normalized deviations, before random-reference ratios, are 0.336 for exact top-k, 0.360 for probe top-k, 0.363 for POKE (top-$k$), 0.973 for DivPrune, and 1.314 for random. These medians need not match the medians of frame-wise ratios. DivPrune lies between random and top-k in this offline measure, as it does in closed loop, which supports the proxy for these rules without validating it for every selection rule.

\begin{table*}[tbp]
    \centering
    \caption{Frame-paired Wilcoxon signed-rank tests (Set 1, $n=120$, \texttt{delete\_k}). Negative differences favor A. These exploratory tests do not account for within-episode dependence and are not used as confirmatory evidence of significance or equivalence.}
    \label{tab:app_wilcoxon}
    \vlaAppendixFit{\textwidth}{
    \begin{tabular}{lccc}
    \toprule
    A $-$ B & Median difference & A win fraction & $p$ \\
    \midrule
    POKE (top-$k$) $-$ Probe($P{=}1$) top-k & $-0.011$ & 0.533 & 0.44 \\
    POKE (top-$k$) $-$ Probe($P{=}3$) top-k & $+0.010$ & 0.458 & 0.13 \\
    POKE (top-$k$) $-$ Exact top-k          & $+0.013$ & 0.358 & $9.8\times10^{-4}$ \\
    Probe($P{=}1$) top-k $-$ Exact top-k   & $+0.021$ & 0.392 & $2.2\times10^{-3}$ \\
    POKE (MMDP) $-$ Probe($P{=}1$) MMDP   & $-0.056$ & 0.692 & $8.1\times10^{-6}$ \\
    \bottomrule
    \end{tabular}
    }
\end{table*}

Deviations under POKE (top-$k$) and probe top-$k$ are of similar magnitude, but $p=0.44$ does not demonstrate equivalence. Modest student--teacher Recall@32 (0.597) can coexist with relatively low action deviation when scores are unevenly concentrated. Training across observations may also reduce probe-specific noise; we did not test this directly.

\paragraph{Comparisons on shared observations.}
Table~\ref{tab:paired} summarizes the Set 3 comparison reported in the main paper; task-wise results are below.

\begin{table}[tbp]
    \centering
    \caption{\textbf{Mask fidelity at shared observations} (OFT Long, $B=32$, 625 frames from Vanilla rollouts). Deviation is normalized by a random mask (lower is better); below random: fraction of frames with a smaller deviation than the random mask. VP's temporal signal is updated along the shared trajectory. This measures mask quality on common reference states, not on each method's own deployment states.}
    \label{tab:paired}
    \setlength{\tabcolsep}{4pt}
    \resizebox{\linewidth}{!}{
    \begin{tabular}{lccc}
    \toprule
    Method & Dev./random $\downarrow$ & Below random $\uparrow$ & Tasks $<$ VP \\
    \midrule
    FastV          & 0.371 & 0.914 & 2/10 \\
    VLA-Pruner     & 0.332 & 0.938 & -- \\
    POKE (top-$k$)  & 0.252 & \textbf{0.981} & \textbf{10/10} \\
    POKE (MMDP)   & \textbf{0.244} & 0.973 & \textbf{10/10} \\
    \bottomrule
    \end{tabular}
    }
\end{table}

\begin{table}[tbp]
    \centering
    \caption{Task-level median normalized deviations on shared Long observations (Set 3, $B=32$; lower is better). The lower block reports two-sided task-level sign tests, with counts of tasks favoring a/b based on the median paired frame-wise difference. This statistic need not equal the difference of the separate task medians in the upper block.}
    \label{tab:app_paired_task}
    \vlaAppendixFit{\linewidth}{
    \begin{tabular}{ccccc}
    \toprule
    Task & FastV & VLA-Pruner & POKE (top-$k$) & POKE (MMDP) \\
    \midrule
    0 & 0.360 & 0.323 & 0.280 & \textbf{0.246} \\
    1 & 0.293 & 0.265 & \textbf{0.216} & 0.240 \\
    2 & 0.432 & 0.314 & 0.311 & \textbf{0.265} \\
    3 & 0.296 & 0.312 & 0.243 & \textbf{0.222} \\
    4 & 0.279 & 0.291 & 0.208 & \textbf{0.200} \\
    5 & 0.483 & 0.389 & 0.257 & \textbf{0.249} \\
    6 & 0.344 & 0.343 & \textbf{0.225} & 0.229 \\
    7 & 0.412 & 0.387 & 0.376 & \textbf{0.285} \\
    8 & 0.445 & 0.410 & 0.268 & \textbf{0.247} \\
    9 & 0.307 & 0.304 & 0.222 & \textbf{0.216} \\
    \midrule
    All (frame median) & 0.371 & 0.332 & 0.252 & \textbf{0.244} \\
    \bottomrule
\end{tabular}
}

    \vspace{0.6em}
    \vlaAppendixFit{0.8\linewidth}{
    \begin{tabular}{lcc}
    \toprule
    Comparison (a $-$ b) & Tasks & Sign-test $p$ \\
    \midrule
    FastV $-$ VLA-Pruner          & 2/8  & 0.109 \\
    FastV $-$ POKE (top-$k$)       & 0/10 & 0.002 \\
    FastV $-$ POKE (MMDP)        & 0/10 & 0.002 \\
    VLA-Pruner $-$ POKE (top-$k$)  & 0/10 & 0.002 \\
    VLA-Pruner $-$ POKE (MMDP)   & 0/10 & 0.002 \\
    POKE (top-$k$) $-$ POKE (MMDP) & 4/6 & 0.754 \\
    \bottomrule
    \end{tabular}
    }
\end{table}

Both POKE variants have lower separate task medians than both attention baselines on all ten tasks. A frame-wise Wilcoxon comparison of FastV and VP gives $p=4.1\times10^{-5}$, whereas the task-level sign test gives $p=0.109$ (eight tasks favor VP). Treating correlated frames as independent overstates the evidence, so the main text uses task-level tests.
The top-k/MMDP task-level comparison ($p=0.754$) does not separate the two variants, which is not the same as showing they are equivalent.

\paragraph{Where offline fidelity and closed-loop success diverge.}
On Goal demonstration frames (Set 1), exact top-k has lower one-step deviation than exact MMDP (median paired difference $-0.115$, win fraction 80\%, exploratory frame-level $p=9.8\times10^{-13}$). On Long reference rollouts (Set 3), the reported task-level test does not separate the two POKE variants. Yet Long closed-loop success favors POKE (MMDP) for all three training seeds. Multi-step success also depends on recovery after deviations, which one-step fidelity does not capture, so the closed-loop gain of MMDP cannot be explained by one-step fidelity alone. We therefore use the offline proxy as a diagnostic next to rollouts, not in place of them.

%% file: main.bbl
\begin{thebibliography}{31}
\providecommand{\natexlab}[1]{#1}
\providecommand{\url}[1]{\texttt{#1}}
\expandafter\ifx\csname urlstyle\endcsname\relax
  \providecommand{\doi}[1]{doi: #1}\else
  \providecommand{\doi}{doi: \begingroup \urlstyle{rm}\Url}\fi

\bibitem[Alvar et~al.(2025)Alvar, Singh, Akbari, and Zhang]{alvar2025divprune}
Saeed~Ranjbar Alvar, Gursimran Singh, Mohammad Akbari, and Yong Zhang.
\newblock Divprune: Diversity-based visual token pruning for large multimodal models.
\newblock In \emph{2025 IEEE/CVF Conference on Computer Vision and Pattern Recognition (CVPR)}, pages 9392--9401. IEEE, 2025.

\bibitem[Black et~al.(2024)Black, Brown, Driess, Esmail, Equi, Finn, Fusai, Groom, Hausman, Ichter, et~al.]{black2024pi0}
Kevin Black, Noah Brown, Danny Driess, Adnan Esmail, Michael Equi, Chelsea Finn, Niccolo Fusai, Lachy Groom, Karol Hausman, Brian Ichter, et~al.
\newblock $\pi_{0}$: A vision-language-action flow model for general robot control.
\newblock \emph{arXiv preprint arXiv:2410.24164}, 2024.

\bibitem[Chen et~al.(2024)Chen, Zhao, Liu, Bai, Lin, Zhou, and Chang]{chen2024image}
Liang Chen, Haozhe Zhao, Tianyu Liu, Shuai Bai, Junyang Lin, Chang Zhou, and Baobao Chang.
\newblock An image is worth 1/2 tokens after layer 2: Plug-and-play inference acceleration for large vision-language models.
\newblock In \emph{European Conference on Computer Vision}, pages 19--35. Springer, 2024.

\bibitem[Fei et~al.(2025)Fei, Wang, Shi, Dai, Cai, Qian, Ji, He, Zhang, Fei, Fu, Gong, and Qiu]{fei2025liberoplus}
Senyu Fei, Siyin Wang, Junhao Shi, Zihao Dai, Jikun Cai, Pengfang Qian, Li Ji, Xinzhe He, Shiduo Zhang, Zhaoye Fei, Jinlan Fu, Jingjing Gong, and Xipeng Qiu.
\newblock Libero-plus: In-depth robustness analysis of vision-language-action models.
\newblock \emph{arXiv preprint arXiv:2510.13626}, 2025.

\bibitem[Intelligence et~al.(2025)Intelligence, Black, Brown, Darpinian, Dhabalia, Driess, Esmail, Equi, Finn, Fusai, et~al.]{intelligence2025pi}
Physical Intelligence, Kevin Black, Noah Brown, James Darpinian, Karan Dhabalia, Danny Driess, Adnan Esmail, Michael Equi, Chelsea Finn, Niccolo Fusai, et~al.
\newblock $\pi_{0.5}$: a vision-language-action model with open-world generalization.
\newblock \emph{arXiv preprint arXiv:2504.16054}, 2025.

\bibitem[Jiang et~al.(2025)Jiang, Jiang, Ma, Wen, Li, Zhan, Jia, Liu, Sun, and Lang]{jiang2025better}
Titong Jiang, Xuefeng Jiang, Yuan Ma, Xin Wen, Bailin Li, Kun Zhan, Peng Jia, Yahui Liu, Sheng Sun, and Xianpeng Lang.
\newblock The better you learn, the smarter you prune: Towards efficient vision-language-action models via differentiable token pruning.
\newblock \emph{arXiv preprint arXiv:2509.12594}, 2025.

\bibitem[Kim et~al.(2024)Kim, Pertsch, Karamcheti, Xiao, Balakrishna, Nair, Rafailov, Foster, Lam, Sanketi, et~al.]{kim2024openvla}
Moo~Jin Kim, Karl Pertsch, Siddharth Karamcheti, Ted Xiao, Ashwin Balakrishna, Suraj Nair, Rafael Rafailov, Ethan Foster, Grace Lam, Pannag Sanketi, et~al.
\newblock Openvla: An open-source vision-language-action model.
\newblock \emph{arXiv preprint arXiv:2406.09246}, 2024.

\bibitem[Kim et~al.(2025)Kim, Finn, and Liang]{kim2025fine}
Moo~Jin Kim, Chelsea Finn, and Percy Liang.
\newblock Fine-tuning vision-language-action models: Optimizing speed and success.
\newblock \emph{arXiv preprint arXiv:2502.19645}, 2025.

\bibitem[Li et~al.(2023)Li, Zhang, Xu, Wang, Yan, Xia, Yang, Cao, Sun, Deng, et~al.]{li2023constraint}
Junyan Li, Li~Lyna Zhang, Jiahang Xu, Yujing Wang, Shaoguang Yan, Yunqing Xia, Yuqing Yang, Ting Cao, Hao Sun, Weiwei Deng, et~al.
\newblock Constraint-aware and ranking-distilled token pruning for efficient transformer inference.
\newblock In \emph{Proceedings of the 29th ACM SIGKDD Conference on knowledge discovery and data mining}, pages 1280--1290, 2023.

\bibitem[Li et~al.(2024)Li, Hsu, Gu, Pertsch, Mees, Walke, Fu, Lunawat, Sieh, Kirmani, Levine, Wu, Finn, Su, Vuong, and Xiao]{li24simpler}
Xuanlin Li, Kyle Hsu, Jiayuan Gu, Karl Pertsch, Oier Mees, Homer~Rich Walke, Chuyuan Fu, Ishikaa Lunawat, Isabel Sieh, Sean Kirmani, Sergey Levine, Jiajun Wu, Chelsea Finn, Hao Su, Quan Vuong, and Ted Xiao.
\newblock Evaluating real-world robot manipulation policies in simulation.
\newblock \emph{arXiv preprint arXiv:2405.05941}, 2024.

\bibitem[Li et~al.(2025)Li, Meng, Sun, Ji, Tang, Fan, Ma, Xia, Wang, and Zhu]{li2025spvla}
Ye Li, Yuan Meng, Zewen Sun, Kangye Ji, Chen Tang, Jiajun Fan, Xinzhu Ma, Shu-Tao Xia, Zhi Wang, and Wenwu Zhu.
\newblock Sp-vla: A joint model scheduling and token pruning approach for vla model acceleration.
\newblock In \emph{The Fourteenth International Conference on Learning Representations}, 2025.

\bibitem[Liang et~al.(2022)Liang, Ge, Tong, Song, Wang, and Xie]{liang2022not}
Youwei Liang, Chongjian Ge, Zhan Tong, Yibing Song, Jue Wang, and Pengtao Xie.
\newblock Not all patches are what you need: Expediting vision transformers via token reorganizations.
\newblock \emph{arXiv preprint arXiv:2202.07800}, 2022.

\bibitem[Liu et~al.(2023)Liu, Zhu, Gao, Feng, Liu, Zhu, and Stone]{liu2023libero}
Bo Liu, Yifeng Zhu, Chongkai Gao, Yihao Feng, Qiang Liu, Yuke Zhu, and Peter Stone.
\newblock Libero: Benchmarking knowledge transfer for lifelong robot learning.
\newblock \emph{Advances in Neural Information Processing Systems}, 36:\penalty0 44776--44791, 2023.

\bibitem[Liu et~al.(2024)Liu, Li, Li, and Lee]{liu2024improved}
Haotian Liu, Chunyuan Li, Yuheng Li, and Yong~Jae Lee.
\newblock Improved baselines with visual instruction tuning.
\newblock In \emph{2024 IEEE/CVF Conference on Computer Vision and Pattern Recognition (CVPR)}, pages 26286--26296. IEEE, 2024.

\bibitem[Liu et~al.(2025)Liu, Chen, Cai, Lin, Yang, Liu, and Zhao]{liu2025bridging}
Ziyan Liu, Yeqiu Chen, Hongyi Cai, Tao Lin, Shuo Yang, Zheng Liu, and Bo Zhao.
\newblock Bridging the semantic-action gap in visual token pruning for efficient vla inference.
\newblock \emph{arXiv preprint arXiv:2511.16449}, 2025.

\bibitem[Molchanov et~al.(2017)Molchanov, Tyree, Karras, Aila, and Kautz]{molchanov2017pruning}
Pavlo Molchanov, Stephen Tyree, Tero Karras, Timo Aila, and Jan Kautz.
\newblock Pruning convolutional neural networks for resource efficient inference.
\newblock In \emph{International Conference on Learning Representations}, 2017.

\bibitem[Pei et~al.(2026)Pei, Chen, Xu, Wang, Shi, and Xu]{pei2026action}
Xiaohuan Pei, Yuxing Chen, Siyu Xu, Yunke Wang, Yuheng Shi, and Chang Xu.
\newblock Action-aware dynamic pruning for efficient vision-language-action manipulation.
\newblock In \emph{International Conference on Learning Representations}, pages 10832--10851, 2026.

\bibitem[Peng et~al.(2005)Peng, Long, and Ding]{peng2005feature}
Hanchuan Peng, Fuhui Long, and Chris Ding.
\newblock Feature selection based on mutual information criteria of max-dependency, max-relevance, and min-redundancy.
\newblock \emph{IEEE Transactions on pattern analysis and machine intelligence}, 27\penalty0 (8):\penalty0 1226--1238, 2005.

\bibitem[Rao et~al.(2021)Rao, Zhao, Liu, Lu, Zhou, and Hsieh]{rao2021dynamicvit}
Yongming Rao, Wenliang Zhao, Benlin Liu, Jiwen Lu, Jie Zhou, and Cho-Jui Hsieh.
\newblock Dynamicvit: Efficient vision transformers with dynamic token sparsification.
\newblock \emph{Advances in neural information processing systems}, 34:\penalty0 13937--13949, 2021.

\bibitem[Ren(2024)]{ren2024openpizero}
Allen~Z. Ren.
\newblock open-pi-zero: Re-implementation of pi0 vision-language-action (vla) model from physical intelligence.
\newblock \url{https://github.com/allenzren/open-pi-zero}, 2024.
\newblock GitHub repository.

\bibitem[Sun et~al.(2026)Sun, Deng, Li, Wang, Geng, and Yu]{sun2026learning}
Yuyao Sun, Tao Deng, Shuang Li, Deqing Wang, Hao Geng, and Minjun Yu.
\newblock Learning to predict middle-layer attention in mllms for visual token pruning.
\newblock \emph{arXiv preprint arXiv:2608.06411}, 2026.

\bibitem[Tao et~al.(2025)Tao, Tang, Wang, Zhu, Hu, and Wang]{tao2025saliency}
Yao Tao, Yehui Tang, Yun Wang, Mingjian Zhu, Hailin Hu, and Yunhe Wang.
\newblock Saliency-driven dynamic token pruning for large language models.
\newblock \emph{arXiv preprint arXiv:2504.04514}, 2025.

\bibitem[Walke et~al.(2023)Walke, Black, Zhao, Vuong, Zheng, Hansen-Estruch, He, Myers, Kim, Du, et~al.]{walke2023bridgedata}
Homer~Rich Walke, Kevin Black, Tony~Z Zhao, Quan Vuong, Chongyi Zheng, Philippe Hansen-Estruch, Andre~Wang He, Vivek Myers, Moo~Jin Kim, Max Du, et~al.
\newblock Bridgedata v2: A dataset for robot learning at scale.
\newblock In \emph{Conference on robot learning}, pages 1723--1736. PMLR, 2023.

\bibitem[Wang et~al.(2025)Wang, Xu, Xiang, Pan, Zhou, Li, and Dai]{wang2025specprune}
Hanzhen Wang, Jiaming Xu, Yushun Xiang, Jiayi Pan, Yongkang Zhou, Yong-Lu Li, and Guohao Dai.
\newblock Specprune-vla: Accelerating vision-language-action models via action-aware self-speculative pruning.
\newblock \emph{arXiv preprint arXiv:2509.05614}, 2025.

\bibitem[Wen et~al.(2025)Wen, Gao, Wang, Zhang, Zhang, Li, He, and Zhang]{wen2025stop}
Zichen Wen, Yifeng Gao, Shaobo Wang, Junyuan Zhang, Qintong Zhang, Weijia Li, Conghui He, and Linfeng Zhang.
\newblock Stop looking for “important tokens” in multimodal language models: Duplication matters more.
\newblock In \emph{Proceedings of the 2025 Conference on Empirical Methods in Natural Language Processing}, pages 9972--9991, 2025.

\bibitem[Xu et~al.(2026)Xu, Wang, Xia, Zhu, Huang, and Xu]{xu2026vla}
Siyu Xu, Yunke Wang, Chenghao Xia, Dihao Zhu, Tao Huang, and Chang Xu.
\newblock Vla-cache: Efficient vision-language-action manipulation via adaptive token caching.
\newblock \emph{Advances in Neural Information Processing Systems}, 38:\penalty0 164448--164473, 2026.

\bibitem[Yang et~al.(2025)Yang, Chen, Tian, Wang, Li, Yu, and Jia]{yang2025visionzip}
Senqiao Yang, Yukang Chen, Zhuotao Tian, Chengyao Wang, Jingyao Li, Bei Yu, and Jiaya Jia.
\newblock Visionzip: Longer is better but not necessary in vision language models.
\newblock In \emph{2025 IEEE/CVF Conference on Computer Vision and Pattern Recognition (CVPR)}, pages 19792--19802. IEEE, 2025.

\bibitem[Yang et~al.(2026)Yang, Wang, Wen, Zhongwei, Zou, Zhang, Wen, and Zhang]{yang2026efficientvla}
Yantai Yang, Yuhao Wang, Zichen Wen, Luo Zhongwei, Chang Zou, Zhipeng Zhang, Chuan Wen, and Linfeng Zhang.
\newblock Efficientvla: Training-free acceleration and compression for vision-language-action models.
\newblock \emph{Advances in Neural Information Processing Systems}, 38:\penalty0 40891--40914, 2026.

\bibitem[Zhang et~al.(2025)Zhang, Cheng, Lu, Zhang, Zhuo, Cao, Guo, She, and Zhang]{zhang2025beyond}
Qizhe Zhang, Aosong Cheng, Ming Lu, Renrui Zhang, Zhiyong Zhuo, Jiajun Cao, Shaobo Guo, Qi She, and Shanghang Zhang.
\newblock Beyond text-visual attention: Exploiting visual cues for effective token pruning in vlms.
\newblock In \emph{Proceedings of the IEEE/CVF International Conference on Computer Vision}, pages 20857--20867, 2025.

\bibitem[Zhang et~al.(2026)Zhang, Liu, Li, Lu, Zhang, Pan, She, and Zhang]{zhang2026beyond}
Qizhe Zhang, Mengzhen Liu, Lichen Li, Ming Lu, Yuan Zhang, Junwen Pan, Qi She, and Shanghang Zhang.
\newblock Beyond attention or similarity: Maximizing conditional diversity for token pruning in mllms.
\newblock \emph{Advances in Neural Information Processing Systems}, 38:\penalty0 25438--25468, 2026.

\bibitem[Zou et~al.(2026)Zou, Lu, Wang, Yan, Lyu, Zheng, Zhang, and Hu]{zou2026don}
Xin Zou, Di Lu, Yizhou Wang, Yibo Yan, Yuanhuiyi Lyu, Xu Zheng, Linfeng Zhang, and Xuming Hu.
\newblock Don't just chase “highlighted tokens” in mllms: Revisiting visual holistic context retention.
\newblock \emph{Advances in Neural Information Processing Systems}, 38:\penalty0 39800--39832, 2026.

\end{thebibliography}
